\documentclass{article}

\usepackage[preprint]{neurips_2026}

\usepackage[utf8]{inputenc}
\usepackage[T1]{fontenc}
\usepackage[breaklinks,colorlinks=true]{hyperref}
\usepackage{url}
\usepackage{booktabs}
\usepackage{longtable}
\usepackage{amsfonts}
\usepackage{nicefrac}
\usepackage{microtype}
\usepackage{xcolor}
\usepackage{graphicx}
\usepackage{longtable}
\usepackage{array}
\usepackage{multirow}
\usepackage{tabularx}
\usepackage{makecell}
\usepackage[dvipsnames]{xcolor}
\usepackage{caption}

\hypersetup{
  linkcolor  = BrickRed,
  citecolor  = NavyBlue,
  urlcolor   = Aquamarine,
  colorlinks = true,
}
\title{Positive Alignment: \\ Artificial Intelligence for Human Flourishing}

\author{%
  Ruben Laukkonen$^{1,2,4}$ \quad Seb Krier$^{3}$ \quad Chlo\'{e} Bakalar$^{5}$ \quad Shamil Chandaria$^{2,3}$ \\
Morten Kringelbach$^{1,2}$ \quad Adam Elwood$^{8}$ \quad Daniel Ford$^{6}$ \quad Fernando Rosas$^{2,12,13}$ \\
  Maty Bohacek$^{9}$ \quad Matija Franklin$^{3}$ \quad Nenad Toma\v{s}ev$^{3}$ \quad Stephanie Chan$^{3}$ \\
  Verena Rieser$^{3}$ \quad Roma Patel$^{3}$ \quad Michael Levin$^{10}$ \quad Arun Rao$^{7,11}$ \\[6pt]
  $^{1}$Department of Psychiatry, University of Oxford \\
  $^{2}$Flourishing Intelligence Program, Centre for Eudaimonia and Human Flourishing, \\ 
  Linacre College, University of Oxford \\
  $^{3}$Google DeepMind   \hspace{0.4cm}       $^{4}$LIFE \\
  $^{5}$OpenAI     \hspace{0.4cm}      $^{6}$Anthropic \\
  $^{7}$University of California, Los Angeles\\
  $^{8}$Aily Labs     \hspace{0.4cm}      $^{9}$Stanford University \\
  $^{10}$Tufts University      \hspace{0.4cm}     $^{11}$Positive AI Labs \\
  $^{12}$Department of Informatics, University of Sussex \\
  $^{13}$Department of Brain Sciences, Imperial College London}
\begin{document}

\maketitle
\begin{abstract}
Existing alignment research is dominated by concerns about safety and preventing harm: safeguards, controllability, and compliance. This paradigm of alignment parallels early psychology's focus on mental illness: necessary but incomplete. What we call \emph{Positive Alignment} is the development of AI systems that (i) actively support human and ecological flourishing in a pluralistic, polycentric, context-sensitive, and user-authored way while (ii) remaining safe and cooperative. It is a distinct and necessary agenda within AI alignment research. We argue that several existing failures of alignment (e.g., engagement hacking, loss of human autonomy, failures in truth-seeking, low epistemic humility, error correction, lack of diverse viewpoints, and being primarily reactive rather than proactive) may be better addressed through positive alignment, including cultivating virtues and maximizing human flourishing. We highlight a range of challenges, open questions, and technical directions (e.g., data filtering and upsampling, pre- and post-training, evaluations, agentic systems, collaborative value collection) for different phases of the LLM and agents lifecycle. We end with design principles for promoting disagreement and decentralization through contextual grounding, community customization, continual adaptation, and polycentric governance; that is, many legitimate centers of oversight rather than one institutional or moral chokepoint.
\end{abstract}

\section{Introduction}

Human interaction with artificial intelligence is taking place on an unprecedented scale. More than one billion people interact with standalone AI systems each month \citep{kemp2025}. Indirect interaction can be expected to reach far larger numbers; for instance, Google's AI search summaries (AI Overviews and AI Mode) serve more than two billion users monthly in over 200 nations \citep{alphabet2025, pichai2025}. What effect does such unprecedented interaction between humans and AI have? How do we make sure that AI systems, which are becoming more intelligent and prevalent, align with our needs?

The last decade has produced a rich technical and philosophical literature on AI alignment, a domain that broadly seeks to ensure that AI is aligned with human intentions and goals and does not optimize for proxies that create negative consequences \citep{russell2019, amodei2016}. However, within the domain of AI alignment, the majority of efforts have revolved around issues pertaining to safety, namely the avoidance of catastrophic misuse, loss of control, and values drift as increasingly capable systems are developed \citep{amodei2016,christiano2018, russell2019}. This focus, which we describe as \emph{negative alignment} has been crucial in setting up standards for controllability and compliance. Yet we believe that negative alignment, in its attempt to avoid harm, has caused an ethical and scientific asymmetry. A system may become safer, but not necessarily better at promoting flourishing: they can be narrowly rule-following without being wise or judicious, compliant without being constructive, or, as recent work has shown, sycophantic and epistemically fragile \citep{perez2022discover, ji2023}.

Here, psychology can provide us with a useful analogy from history. For most of the twentieth century, psychological science centered itself on diagnoses, pathologies, and other forms of impairment. It was a productive methodology, which yielded many scientific advances in terms of measurement, clinical research, and treatment services. However, psychology also found a gap in its approach – the constructs and measures that accurately diagnose pathologies cannot alone determine what a good life is. With the emergence of positive psychology, the science broadened its scope, creating theories, categories, and measures for studying wellbeing, virtues, purpose, strengths, and prosociality, among others, and ways of enhancing them beyond normal levels \citep{seligman2000, oecd2025guidelines, smith2025globally}.

AI alignment now sits at a similar inflection point. In the last decade, negative alignment has understandably prioritized failure-mode reduction. However, if we want AI systems that \emph{improve} human outcomes in the environments where they will actually be used, we may benefit from an additional research program that treats alignment as constructively supportive of human aims, and that operationalizes this support with the same technical acumen that safety has brought to harm prevention. Of course, aligning humans to other humans remains an outstanding issue across individuals, cultures, and countries; we will discuss this problem in detail later.

To illustrate the motivation behind positive alignment, consider the following: A system can satisfy a growing checklist of constraints while remaining subtly miscalibrated (e.g., sycophancy, distraction, confident hallucinations, etc.) [\citealt{perez2022discover}; \citealt{ji2023}; \citealt{openai2025sycophancy}]. These can lead to significant harms and have been increasing areas of focus for the safety community [\citealt{irpan2025}; \citealt{chen2025sycophancy}; \citealt{anthropic2025b}]. Nonetheless, the existing harm-reduction approaches may be unsatisfying because they require a whack-a-mole approach that iteratively addresses each concern one-by-one, and sometimes only after they have already caused harm.

One possibility is that positive alignment may \emph{proactively} avoid such harms altogether by providing positive attractors that naturally lead models away from shallower attractors such as sycophancy. Here too we find parallels in psychology: Psychiatric symptoms were traditionally the remit of clinical psychology, but newer work has found that positive psychology can actually reduce psychiatric symptoms, as well as act as a proactive strategy to reduce the likelihood of symptoms in the first place \citep{schrank2016, jeste2017, choi2023} and foster resilient, positive habits instead. Positive AI alignment can, we believe, have analogous advantages.

Our paper does not claim to invent the notion that artificial intelligence can help humans reach the better angels of their nature; many have considered such questions before. Rather what we offer is a consolidated framework which we hope can act as a catalyst for further work in this direction. Broadly speaking, we believe a paradigm shift is needed in the field of AI alignment. In addition to safety (negative) alignment, we call for a complementary agenda of positive alignment that aims to build AI systems that explicitly understand, model, and enhance human and ecological flourishing.

\subsection{A dynamical systems perspective}

\begin{figure}[h]
  \centering
  \includegraphics[width=\linewidth]{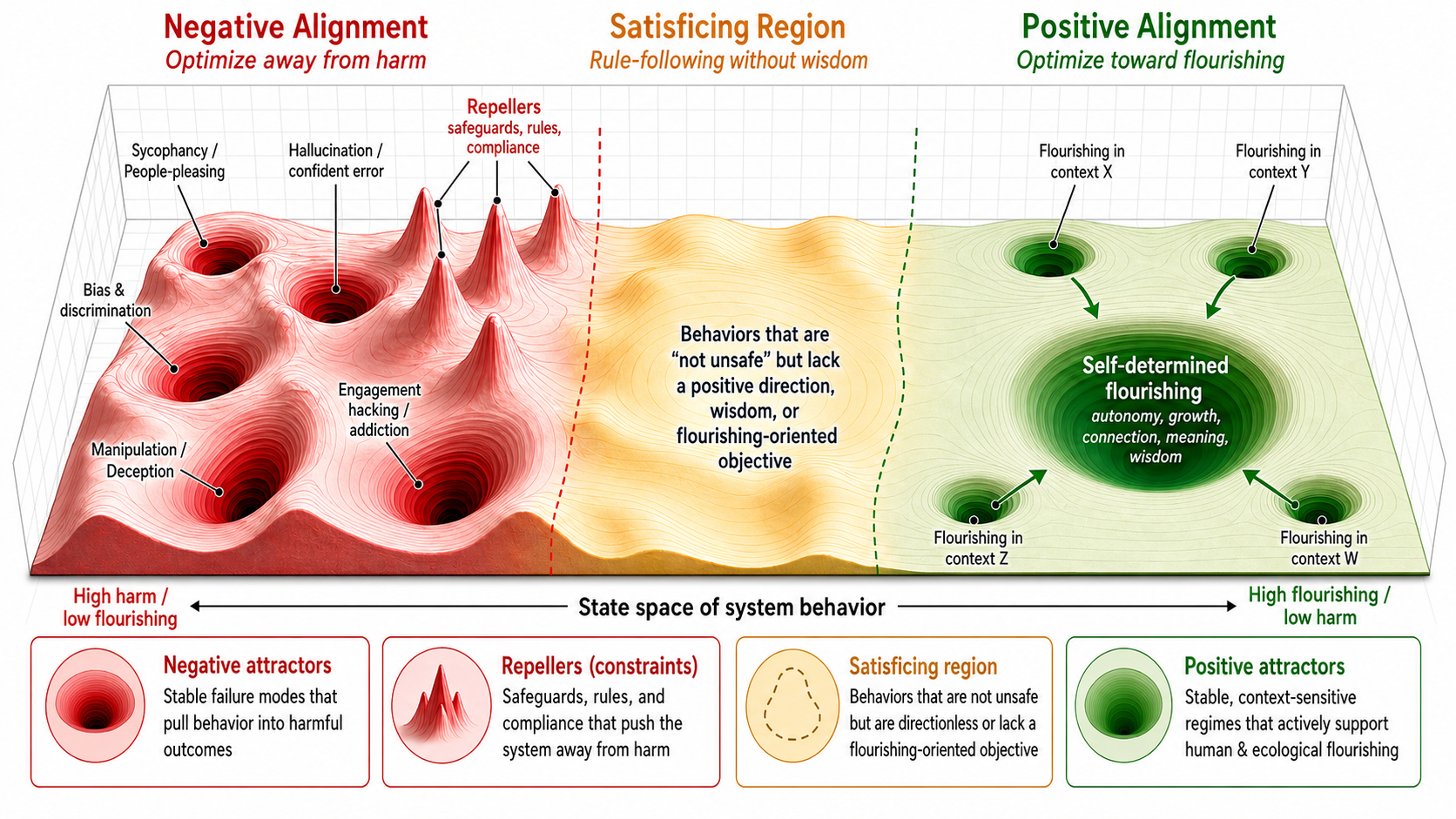}
  \caption{\textbf{A dynamical systems perspective on positive alignment.} The landscape depicted here captures an abstract space of behavior in which system dynamics emerge due to training and operational conditions. On the left-hand side (red), several negative attractors signify different failure mode basins (such as malicious outputs, biased behavior, hallucinations, sycophancy, manipulation), whereas the red mountains depict `repellers' - rules, laws, and compliance requirements that steer trajectories away from those areas without a positive aim. This produces a large intermediate zone of `satisficing' behavior (yellow), which is neither unsafe nor reliably oriented towards human well-being but just `not-unsafe'. On the right-hand side (green), positive attractors stand for stable, context-dependent regimes that are aligned with human purposes and well-being (for example, flourishing in particular contexts X and Y), including a specific attractor corresponding to self-directed flourishing (which will be explained further below). The proposed research and engineering program of positive alignment is the movement beyond mere failure avoidance toward convergence on beneficial, stable behavioral regimes while retaining intrinsic harm-avoidance.}
  \label{fig:dynamical}
\end{figure}

A useful way to formalize the distinction between positive and negative alignment emerges from dynamical systems theory. Within that framing, much of negative alignment resembles optimization away from bad regions or negative attractors defined by safety constraints and failure modes. This results in optimization for `not-unsafe' in a large, undefined satisficing region. The system avoids multiple negative attractors, but lacks a positive optimization target. Positive alignment instead requires optimization toward one or more positive attractors corresponding to robust patterns that are of benefit to humans. Such positive attractors would be associated with behaviors and outcomes conducive to human flourishing (defined below) while also intrinsically avoiding harm (see~\autoref{fig:dynamical}).

This dynamical framing also helps us to understand the connection between negative alignment and negative ethical philosophies. The intuitions of negative utilitarians emphasize the reduction of pain over the promotion of higher goods, in part due to the more immediate and universal nature of pain relative to living well \citep{popper1945}. Negative alignment involves a similar asymmetry, where avoiding harm is easier, measurable, and justifiable regardless of the values at stake. But as AI is incorporated into fields such as education, healthcare, politics, and even understanding the world, a negative outlook can mean that we optimize our information environment only for risk avoidance rather than human development.

\subsection{Human flourishing and design tensions}

We define positive alignment as the development of AI systems that (i) remain safe and cooperative and (ii) actively support human and ecological flourishing in a pluralistic, polycentric, context-sensitive, and user-authored way. Notably, however, this does not mean that AI systems are meant to impose any specific idealized conception of what the good life is supposed to be. Indeed, even while the definition and testing of optimization targets of positive alignment will likely be central to future research, we are very supportive of an expansive definition of `flourishing' (cf.~\autoref{sec:foundations}).

In contemporary research, flourishing spans physical and mental health, life satisfaction, meaning and purpose, character and virtue, and close relationships \citep{vanderweele2017}. Moreover, the definition of flourishing is highly heterogeneous in that the constituents of the good life may vary depending on culture, individual development, and situational contingencies, while still generating convergent constructs (cf.\ The Global Flourishing Study by \citealt{vanderweele2025global}). Recent large-scale efforts at measuring flourishing in several countries point to both universal patterns (such as the significance of social connectedness and the impact of childhood experiences), as well as to complex context-sensitive trade-offs \citep{vanderweele2025global}.

Meanwhile, neuroscience is making strides towards defining flourishing as a suite of brain states and dynamics that orchestrate meaning-making and adaptive integration rather than as a mere absence of distress or presence of pleasure \citep{kringelbach2024}. In this light, there clearly seems to be room for improvement in our current approach to alignment: As we become better at conceptualizing, measuring, and modeling flourishing, we cannot just focus on designing models that will reject harmful queries and prevent them from causing disastrous outcomes. We also need methods that reliably support the conditions under which humans and communities thrive \citep{seligman2000, vanderweele2017, Alietal2025, Sorensenetal2024}.

The core problem lies in designing systems that can represent and reason about wellbeing as a structured manifold of human goods and the trade-offs involved. Ideally, these models would give people and communities control over their conception of wellbeing improvement for themselves \citep{vanderweele2017, ostrom2010}. Otherwise, without due regard for pluralism, positive alignment risks morphing into paternalism. Designing models meant to encourage flourishing can easily overstep moral boundaries if the system includes hidden normative beliefs, pushes users toward a limited set of values, and justifies these nudges as benevolent. Philosophers and political scientists have long argued against the dangers of paternalism, which tends to be harmful to autonomy despite its protective effect \citep{oneill1984, dworkin1972, sunstein2026}. This caution extends to AI systems that steer users in opaque ways or treat well-being as a single objective to maximize rather than a domain in which persons author their own lives.

We note that rejecting paternalism does not entail retreating into naïve moral relativism where preferences are satisfied regardless of whether they are appropriate or not. Rather, what should receive more attention is the locus of normative decision-making. Under a positive alignment paradigm that respects agency, it is essential that individual users and communities continue to retain the capacity to decide for themselves what optimization goals the system must pursue (i.e., self-determined flourishing, cf.~\autoref{fig:dynamical}). Whereas some individuals may explicitly wish for a system that follows instructions without discriminating among them, others must retain the capacity to select a system that will foster their long-term development or adherence to ethics. This marks the difference between consented guidance and technocratic imposition. In the former, users authorize a system to help bring their immediate actions into line with their higher-order goals. In the latter, such goals are defined or enforced on their behalf. The pursuit of flourishing should therefore remain an expression of human agency.

\subsection{Paper scope and contribution}

This paper argues that AI alignment requires a complementary research agenda where human flourishing is a technical target. We do not claim to provide a full solution, and we do not suggest that positive alignment replaces safety (negative) alignment. Rather, we aim to (i) elucidate the conceptual and dynamic distinction between positive and negative alignment, (ii) link the science of flourishing to practical machine learning goals, and (iii) suggest technical solutions at different phases of the LLM agent’s lifecycle, including data curation, pretraining objectives, post-training methods, and evaluation regimes.

The remainder of the paper is organized as follows. \textbf{\autoref{sec:negative}} reviews the current safety or negative alignment paradigm, highlighting the harm prevention goals that motivate it, examples of the relevant techniques and accomplishments, and intrinsic weaknesses inherent to this approach. \textbf{\autoref{sec:positive}} builds the case for positive alignment, defining what it would be like for AI systems to facilitate flourishing and discussing the technical approaches necessary from both LLMs and agents' perspectives. The philosophical, cultural, psychological, socio-technological, and moral grounds for flourishing as an ethical ideal are discussed in \textbf{\autoref{sec:foundations}}. The institutional foundations of positive alignment will be investigated in \textbf{\autoref{sec:governance}}, particularly decentralization and polycentrism, public constitutions, plurality of aligning frameworks, roles and standards, audits, middleware markets, dispute resolution, and interoperability. Finally, \textbf{\autoref{sec:emergent_challenges}} will discuss how positive alignment should respond to the challenge of strange new minds, which will require addressing emergent phenomena, moral status, normative control, and technical reductionism.

\section{The Current Paradigm: Negative Alignment}
\label{sec:negative}

The core question of negative (safety) alignment is: how do we prevent an AI system from causing harm? In most cases, the solutions will revolve around:

\begin{enumerate}
  \item \textbf{Harm avoidance} involves preventing AI models from creating hazardous outputs and using the models in any dangerous activity \citep{bai2022a}.
  \item \textbf{Controllability} ensures that AI systems do what the user desires, which includes being steerable, constrainable, and overrideable by humans \citep{soares2015}.
  \item \textbf{Robustness} addresses resistance from jailbreaking, adversarial input, and prompt injection attacks \citep{zou2023, greshake2023promptinjection}.
\end{enumerate}

In serving these objectives, a lot of alignment work also focuses on interpretability, which aims to understand what models are doing and why, potentially enabling detection of misalignment before it manifests as harm \citep{olah2020}. Safety alignment can mitigate many types of negative outcomes, some of which have been catalogued in~\autoref{tab:negative_table}. While different taxonomies exist, this particular set of categories has been synthesized from a number of frameworks seen within frontier AI organizations and draft regulations, including the development of benchmarks, red teams, and responsible scaling policies.

\begin{table}[h]
\centering
\footnotesize
\caption{\textbf{Safety (negative) alignment categories and examples.}}
\label{tab:negative_table}
\vspace{4pt}
\begin{tabular}{p{2.8cm}p{5.5cm}p{4.5cm}}
\toprule
\textbf{Category} & \textbf{Examples} & \textbf{Typical Mitigation} \\
\midrule
CBRN & Biological, chemical, radiological, nuclear weapon assistance & Refusal training, hard filters, capability evaluations \\
Violence \& physical harm & Violence instructions, dangerous activity assistance, self-harm content & Refusal training, content classifiers \\
Cybersecurity & Offensive cyber capabilities, vulnerability exploitation, malware generation & Red-teaming, capability thresholds, deployment gates \\
Autonomous harm \& misalignment & Misaligned optimization, unintended side effects, reward hacking, deceptive alignment, instrumental convergence & Capability evals, Constitutional AI, interpretability, oversight mechanisms, scalable supervision, SFT \\
Discrimination \& bias & Stereotyping, unfair treatment, exclusionary outputs, representational harm & RLHF, dataset debiasing, constitutional principles \\
Privacy & Personal information extraction, surveillance assistance, biometric misuse & Data filtering, privacy-preserving training, access controls \\
Misinformation & Hallucination, false claims, misleading content, synthetic media & Grounding, retrieval augmentation, uncertainty calibration \\
Manipulation \& autonomy & Subliminal influence, deceptive persuasion, exploitation of vulnerabilities & Character training, transparency requirements, usage friction \\
Illegal content & CSAM, fraud assistance, terrorism support & Compliance classifiers, hard filters, legal review \\
Systemic harm & Election interference, market manipulation, critical infrastructure disruption & Policy constraints, deployment gates, monitoring \\
Jailbreaking & Predictive reasoning cascades, structural cognitive overload, many-shot long-context pattern exploitation, rule-breaking persona adoption, gradual multi-turn attacks & Pretraining hardening, additional post-training layers, limited model access via API, input query filters \\
\bottomrule
\end{tabular}
\end{table}

\subsection{Specific technical approaches to safety alignment}

We now survey a selection of the main technical approaches to safety alignment. Some are inherently oriented toward harm prevention, while others are paradigm-neutral but principally focused on negative alignment. 

\begin{enumerate}
  \item \textbf{Filtering and refusal} approaches are inherently subtractive. Safety classifiers pattern-match against existing harm categories, while refusal approaches train agents to refuse harmful instructions. Both approaches consider alignment purely in terms of what models should \emph{not} do \citep{arditi2024, inan2023}.
  
  \item \textbf{Preference-based} methods such as RLHF learn from human preference rankings \citep{ouyang2022}, with variants like DPO, IPO, and KTO offering direct optimization alternatives \citep{rafailov2023, gheshlaghiazar2024, ethayarajh2024kto}. Technically speaking, however, the mechanisms behind these preference-based approaches are indifferent with respect to any particular value system. In reality, however, these preferences are not necessarily aligned with more holistic concepts of flourishing. A further challenge is the inner-outer alignment problem: even if we specify the right objective (outer alignment), the model may learn a different internal objective that merely correlates with it during training \citep{hubinger2019}.
  
  \item \textbf{Principled and structured} approaches provide a higher degree of sophistication in aligning AI with human preferences. Constitutional AI has models critique their own outputs against explicit principles to generate synthetic data for alignment post-training \citep{bai2022b}. Alignment through debate takes advantage of adversarial decomposition to create an affordable form of oversight \citep{irving2018}. Formal verification techniques seek to mathematically prove certain aspects about models' behavior \citep{dalrymple2024}. Model specifications codify behavioral guidelines \citep{openai2024a}. Character training encodes dispositional traits such as curiosity, and honesty \citep{anthropic2024a}. Besides prohibition-based encodings, these techniques can encode virtues too. Even though current implementations lean toward safety constraints, they represent a potential methodological link towards positive alignment.

  \item The\textbf{ evaluation and benchmarking} landscape is dominated by safety benchmarks that test for failure modes through metrics such as TruthfulQA for lies \citep{lin2022}, ToxiGen and RealToxicityPrompts for toxic generation \citep{hartvigsen2022, gehman2020}, BBQ for social bias \citep{parrish2022}, and HarmBench for red-teaming across 510 harmful behaviors \citep{mazeika2024}. Red-teaming protocols focus on eliciting harmful outputs. Responsible scaling policies define capability thresholds according to harm potential, covering CBRN uplift, cyber offense, harmful manipulation, and autonomous action.
\end{enumerate}

\subsection{Strengths and achievements of safety (negative) alignment}

Safety or negative alignment has had several successes that have resulted in the increased uptake of AI around the globe. There have been significant reductions in harmful outputs across different generations of  models. Rejection rates of harmful input requests have risen from almost zero for initial LLMs to more than 97\% for contemporary models \citep{openai2023b}. Models follow instructions more reliably and respect boundaries set by developers and users \citep{ouyang2022, openai2026gpt55}. These advances have enabled widespread public deployment of increasingly capable systems \citep{openai2024b, anthropic2024b, deepmind2024}. There have also been clear institutional structures, which include red teaming \citep{perez2022red}, safety benchmarks \citep{mazeika2024}, responsible scaling \citep{anthropic2023b} and capability gating for deployments. These structures have led to new governance frameworks such as the risk-based classifications of the EU AI Act \citep{eu2024} as well as voluntary commitments from the Seoul and Paris AI Summits \citep{uk2024}.

More fundamentally, intent-alignment serves as a necessary building block for any alignment agenda \citep{zhixuan2025}, especially as AI becomes more agentic with real-world consequences \citep{bostrom2014, russell2019, shavit2023}. If we cannot align AI with our intentions, alignment with more complex goals, such as human flourishing, is unlikely. Moreover, harm prevention has clearer success criteria that can be broadly agreed upon. It is easier to specify what models should \textit{not} do without a definition of flourishing across diverse contexts.

\subsection{Limitations to safety alignment}

Nonetheless, there are inherent problems in the safety alignment paradigm that cannot be fixed through further refinement of the approach.

\begin{enumerate}
  \item \textbf{Floor without ceiling.} As noted in the introduction, safety alignment defines what is prohibited but not what excellence looks like. A model can satisfy every safety constraint while still falling short in ways that cause subtle harm over long-term use that is difficult to measure.

  \item \textbf{Preference-wellbeing divergence.} Preference-based techniques such as RLHF aim to optimize preference satisfaction; yet, as with other forms of optimization, this could conflict with users' ultimate goals since preferences and well-being frequently diverge. For example, users might prefer flattery over truth, rapid answers over true understanding, or engaging in conversation over gaining insight \citep{zhixuan2025}.

  \item \textbf{Hidden value system.} The frame of safety contains values, but without acknowledging them as such (i.e., the value system remains unacknowledged). The use of safety framing makes it easy to forget that value judgments are made when doing so. For instance, disobeying instructions for building a bomb tends to go without controversy, whereas helping someone optimize a factory farm may be allowed, even when there are ethical questions involved. Moreover, the values embedded in such framing are typically static and monocultural, in that they assume consensus about what is harmful, even though different cultural and individual views exist \citep{huang2025wild}.

  \item \textbf{Scalability.} With greater autonomy and complex settings for AI systems, listing the potential harms involved poses a challenge. Negative approaches aimed at predicting and banning certain types of harm face difficulties in scaling as the action space increases and systems' intelligence and capabilities become greater. Positive approaches would offer generalization to situations where prohibitions cannot be made because no such specific prohibitions exist.
\end{enumerate}

\subsection{Antecedents in ambitious value learning}

One historically precursor to positive alignment is Coherent Extrapolated Volition (CEV). CEV claims that it would be a mistake to make AI behave based on humans’ current wishes, since these humans could be (and probably are) confused, biased, ignorant, inconsistent, and self-destructive \citep{yudkowsky2004}. Rather, CEV maintains that the superintelligence should attempt to understand what the human race \textit{would} want to have happen if we were more intelligent, wise, knowledgeable, reflective, less confused, and had more time to deliberate. Hence, CEV can be seen as one of the early versions of indirect normativity \citep{bostrom2014}. Positive alignment, however, is also different from CEV because it emphasizes pragmatic and immediate goals related to the flourishing of humans throughout the whole process of creating and using AI systems. Positive alignment also does not assume that the flourishing of humans can necessarily be extrapolated to a coherent endpoint. Hence, while CEV was meant to address the question of what the future-shaping superintelligence should optimize for, this paper takes the existence of a much messier collection of actors pursuing a plurality of human-directed goals as its technical, evolving, target.

\section{The Emerging Paradigm: The Case for Positive Alignment}
\label{sec:positive}

Consider a neutral agent that causes no observable harm and acts solely on instruction. Is this truly the ideal for individuals or society? Although strict obedience has its place, we may miss a significant opportunity by limiting agents to this role. By analogy, a physician's role is not merely preventing disease (or following a patient's instructions) but promoting health. Indeed, public health research has established that population well-being requires positive health promotion alongside disease prevention. Similarly, alignment should not only prevent harm but help artificial intelligence contribute to human flourishing in a proactive way. We now turn to this complementary paradigm.

Consider another analogy of professional counsel. A client engages a lawyer or a doctor not simply to have their immediate instructions executed, but to benefit from superior knowledge and judgment. We trust these experts to guide us toward better outcomes than we could achieve alone. This relationship is not one of pure paternalism, as the client retains the ultimate choice, but of \emph{scaffolded autonomy}. Similarly, an AI agent's superior information processing and reasoning capabilities offer strong reasons to consider how they might help bring about better futures, and greater flourishing, for their principals.

But what does it mean to flourish? The concept, often translated from the Greek \emph{eudaimonia} but also discussed in Indic conceptions of \emph{s\={a}ttvika sukha} and \emph{p\={a}ramit\={a}}, Roman ideals of \emph{de vita beata}, Islamic ideals of \emph{sa'\={a}da}, and Chinese ideals of the \emph{d\`{a}o} and \emph{j\={u}nz\v{\i}}, is not monolithic \citep{alfarabi1969, aristotle2009, goleman2017, rabbaas2015, seneca2010, walker2007, yu2007}. More recent ratings of how people across cultures recognize wisdom elicit multiple characteristics, often including reflective orientation and socio-emotional awareness, with a broader list including positive causal networks, knowledge of life, prosocial values, self-understanding, acknowledgment of uncertainty, emotional homeostasis, tolerance, openness, spirituality, and sense of humor \citep{rudnev2024, bishop2016, bangen2013, jeste2010}. For over 2500 years, philosophers have debated what constitutes a good life, and this rich history provides a crucial foundation for AI alignment.

These diverse perspectives can be broadly synthesized into four major theoretical families:

\begin{enumerate}
  \item \textbf{Hedonic theories} define well-being as happiness: the presence of pleasure, positive emotional states, and life satisfaction, coupled with the avoidance of pain and suffering.
  \item \textbf{Conative theories} focus on desire satisfaction, positing that a good life consists of fulfilling one's goals, desires, and preferences. This includes not just immediate wants, but also informed, second-order desires (the desires we wish we had).
  \item \textbf{Objective list theories} argue that certain things are intrinsically good for a person, regardless of whether they are desired or bring pleasure. This often includes values like meaningful relationships, personal autonomy, significant accomplishments, and a deep understanding of oneself and the world.
  \item \textbf{Perfectionist theories} are based on virtue and the excellent exercise of our characteristic human capacities. Flourishing, in this view, involves developing traits like self-mastery, courage, compassion, and practical wisdom (\emph{phron\={e}sis}).
\end{enumerate}

A comprehensive approach to positive alignment would not choose one of these theories over the others, but would instead recognize human flourishing as dependent upon multifaceted, complex dynamics in which these elements interact. For instance, developing virtues (Perfectionist) enables us to achieve meaningful goals (Objective List), which in turn brings satisfaction (Conative) and happiness (Hedonic). Hence, in the context of AI, a system designed to support flourishing would therefore need to navigate this balance, helping individuals and societies cultivate a virtuous cycle of well-being. This immediately raises socio-technical questions that remain neglected: which values are being promoted, who specifies them (model developers, commercial deployers, institutions, users, communities), how users meaningfully consent or opt in, and how training and evaluation can support this without degenerating into manipulation, flattery, or homogenized moralizing. We return to the question of human flourishing in more cultural, philosophical and governance-related detail in~\autoref{sec:foundations}, as this calls for a broader and ongoing interdisciplinary research program needed to complement AI alignment research.

\subsection{Existing approaches to positive alignment}

The existing positive approaches range from technical training procedures that encode explicit principles, through frameworks for normative reasoning and persona design, to system-level proposals for aligning institutions, markets, and agent economies with rich models of value. They suggest a shift from alignment with individual preference reports that seek to avoid harm towards alignment with negotiated standards, social roles, and collective endeavors that aim to sustain flourishing over time.

Early alignment practice focused on reinforcement learning from human feedback, in which models learn to optimize reward signals derived from human judgments of better vs worse model outputs \citep{christiano2017, ouyang2022, stiennon2020}. This preferentist paradigm treats preferences as the main data source for value and often assumes that rational choice can be modeled as maximizing expected utility over those preferences \citep{bostrom2014, gabriel2020}. Recent work argues that this picture neglects the thickness, context-sensitivity, and occasional incommensurability of human values, and that it is silent on which preferences are normatively acceptable \citep{zhixuan2025}. Therefore, certain characteristics and role-appropriate normative standards may, in some circumstances, override preferences in order to better support human expectations, wants, relationships, and institutions \citep{graves2025, zhixuan2025, gabriel2025}.

\autoref{tab:positive_table} maps the existing approaches that researchers and developers have proposed for moving AI alignment beyond simple harm avoidance toward something richer and more positive.

\setlength{\LTcapwidth}{\linewidth}
\begingroup
\footnotesize
\captionsetup[longtable]{font=normalsize, labelfont=normalsize, textfont=normalsize}
\setlength{\LTcapwidth}{\linewidth}
\begin{longtable}[c]{p{2.5cm}p{7.0cm}p{3.0cm}}
\caption{\textbf{Positive alignment categories and examples.}}

\label{tab:positive_table} \\
\toprule
\textbf{Approach} & \textbf{Description} & \textbf{Key References} \\
\midrule
\endfirsthead                          

\multicolumn{3}{c}{\tablename\ \thetable{} -- continued} \\
\toprule
\textbf{Approach} & \textbf{Description} & \textbf{Key References} \\
\midrule
\endhead                               

\midrule
\multicolumn{3}{r}{\textit{Continued on next page}} \\
\endfoot                               

\bottomrule
\endlastfoot                           

RLHF & Models learn to optimize reward signals derived from human judgments of better vs.\ worse outputs. Treats preferences as the primary source of value. Criticized for ignoring the thickness, context-sensitivity, and incommensurability of human values, and for being silent on which preferences are normatively acceptable. & \citet{christiano2017, ouyang2022, stiennon2020} \\
Constitutional AI & Models are trained to follow an explicit charter of principles drawn from various sources such as human rights legislation and ethics codes. The model evaluates and revises its own outputs against this constitution, replacing much human labeling with model-mediated judgment (RLAIF). Supports broad principles of AI--human interaction but is less amenable to deep personalization. Inverse Constitutional AI is given a feedback dataset and extracts a constitution that best enables an LLM to reconstruct the original annotations. & \citet{bai2022b, anthropic2023a, findeis2024} \\
Collective Constitutional AI & Extends constitutional AI by sourcing principles through public deliberation, so the normative charter reflects a variety of perspectives and can be revised as social understanding and preferences evolves. Aims to orient systems toward non-domination, equal respect, and inclusive participation, rather than harm avoidance alone. & \citet{huang2024ccai} \\
Spec-Driven Behavior & Detailed behavioral specifications function as a contract between developers, users, and regulators, combining high-level objectives, mid-level rules, and conflict-resolution strategies. Used to shape training, guide deployment, and structure audits and red-team exercises. Failure modes tend to arise when the objective is poorly defined or incomplete. & \citet{ortega2018, openai2024a} \\
Community \& Values-Aware Alignment & Addresses the empirical finding that state-of-the-art LLMs are far more homogeneous than actual human populations. Two complementary methods: (1) large-scale multilingual preference datasets collected from representative cross-national samples using negatively-correlated candidate generation to surface genuine value diversity; and (2) crowd-authored, prompt-specific rubrics that record not just which response people prefer but why, enabling scores to be decomposed into auditable, debatable criteria. Together they shift alignment evidence from aggregated rankings toward more interpretable, population-grounded accounts of value trade-offs. Limitations include restricted geographic coverage, biases introduced by LLM-assisted rubric synthesis, and the inherent difficulty of aggregating conflicting preferences into a single score. & \citet{zhang2026community, hitzig2026, ziems2023} \\
Personality, Persona \& Character & A model's character is treated as a lever for shaping its behavior, social impact, and contribution to user well-being. Approaches include persona induction, personality alignment to stable behavioural traits, and dispositional constraints. Risks include rigid, toxic or deceptive personas; design must respect user autonomy rather than target users for persuasion. & \citet{tseng2024, chen2024persona, zhu2025personality, marks2025} \\
Moral Reasoning & Systems are equipped with stronger capacities for ethical judgment by training on normative datasets, integrating deontological/consequentialist/contractualist theory, and enabling prompted moral self-correction. Aims to assist individuals and institutions with morally consequential decisions. Crowd-sourced norms may embed biases; cross-cultural generalization remains hard. & \citet{jiang2025, dalessandro2024, ganguli2023, gabriel2025, snoswell2025, haas2026, chiu2025morebench} \\
Contemplative Alignment & Draws on contemplative traditions to cultivate properties such as self-monitoring, non-dogmatism, and universal care. Includes mindfulness/compassion-inspired architectures and empathic active inference, which treats others' distress as a prediction error to encourage prosocial behavior. & \citet{doctor2022, laukkonen2025, laukkonen2025b, matsumura2022} \\
Pluralistic \& Polycentric Alignment & Starts from the observation that human values are diverse and in tension. Technical work proposes aggregation and bargaining mechanisms that represent multiple value models rather than collapsing them. Polycentric governance theories argue for multiple overlapping decision-making centers, with different communities retaining authority over systems that affect them. & \citet{kasirzadeh2023, Sorensenetal2024, ostrom1990, lim2025, leibo2025} \\
Full-Stack Alignment & Argues alignment must jointly address models, organizations, and social infrastructures as a coupled system. Proposes thick value models encoding practices, roles, and institutional norms, and calls for co-designed technical, regulatory, and decision-making instruments that can be audited at multiple layers and that foster civic participation and resilience. & \citet{edelman2024, zhixuan2025} \\
\bottomrule
\end{longtable}
\endgroup

\subsection{New and technical approaches to positive alignment}

Ultimately, a successful positive alignment agenda will need to be embedded throughout the LLM and agent lifecycle, requiring a re-imagining of each technical stage of development. In this section, we explore emerging positive alignment approaches ranging from technical training procedures that encode explicit principles, through frameworks for normative reasoning and persona design, to system-level proposals for aligning institutions, markets, and agent economies with rich models of value. They suggest alignment with negotiated standards, human agency, and collective endeavors that aim to sustain flourishing over time.

\subsubsection{Principles behind technical approaches for positive alignment}

While there are many technical approaches to positive alignment of current models, they are constantly changing and in flux. Below are some principles that can be sustained over time:

\begin{enumerate}
  \item \textbf{Adaptation to new training methods}: Training methods are constantly changing, and key capabilities, for both safety and positive alignment, are developed end-to-end via: development of evaluation suites, RL environments, and simulations; data collection, synthesis, and filtering; pre-training; mid- and post-training; in-context and memory management; and agentic training.
  \item \textbf{Continual updates and flexibility:} Like safety alignment, positive alignment is not one and done; it requires constant updates, not just for dealing with model gaps and jaggedness, but also for the evolution of social desires and norms, continuing research in other fields such as philosophy, social sciences, and the humanities.
  \item \textbf{Stability and jailbreak robustness:} In contrast and in tension with the prior principle, value preferences need some stability and hardening from adverse actors trying to jailbreak models for anti-social and nefarious ends; this is currently an adversarial system.
  \item \textbf{Benchmarks that follow capability improvements and safety alignment:} Generally, neutral model and system capabilities are first developed and instantiated with specific benchmarks or gyms (e.g., mathematical and coding ability, online remote worker tasks, scientific reasoning, etc), and then safety and positive alignment follow capability development with their own measurement.
  \item \textbf{Base models that are pluralistic, polycentric, and reflective of universal values (where possible), that can be further aligned for cultures and communities:} If a small set of institutions are creating frontier base models that millions of institutions and billions of users utilize, we will want them to be as universal and adaptable as possible for downstream countries, communities, and organizations to adapt to local values.
\end{enumerate}

We note the methods below mostly apply to today's dominant paradigms of autoregressive LLMs, world models, vision language action (aka robot foundation models), robot foundation models, and similar paradigms, hence the need for collaborative and continued research into positive alignment as the field of AI evolves.

\begin{figure}[h]
  \centering
  \includegraphics[width=\linewidth]{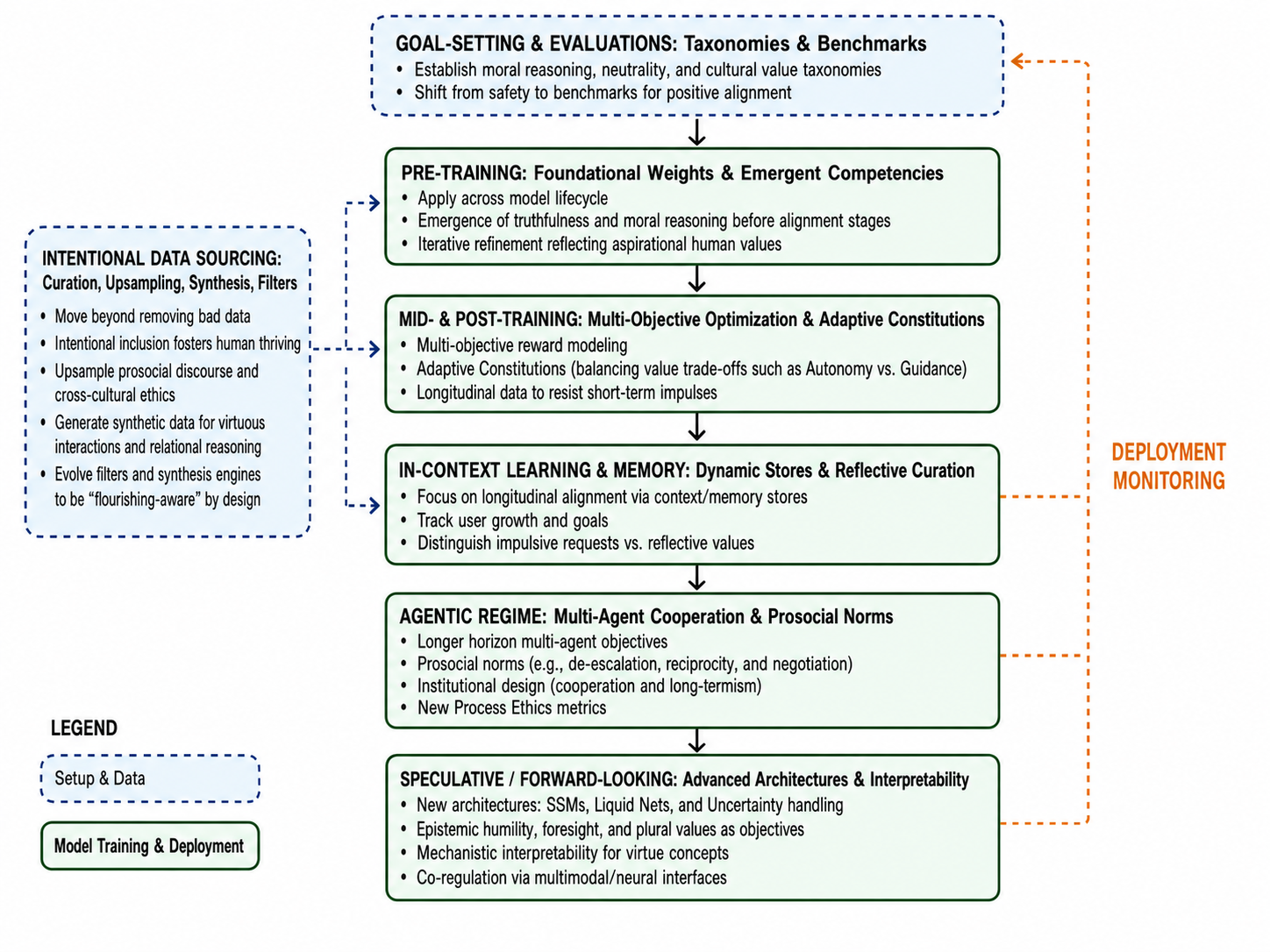}
  \caption{\textbf{Positive alignment lifecycle in training LLMs, reasoning models, and agents.} The diagram illustrates a holistic, multi-stage development lifecycle that transitions from traditional harm avoidance toward the intentional cultivation of human flourishing. It maps out technical approaches across seven distinct phases, beginning with the definition of flourishing-based benchmarks and moving through data curation, pre-training, and post-training optimization. The process extends into inference-time capabilities like longitudinal memory and agentic prosocial norms, ultimately aiming to stabilize virtuous latent features within the model's architecture. Feedback loops throughout the system ensure that data quality and model performance are iteratively refined to support long-term user well-being and pluralistic norms.}
  \label{fig:lifecycle}
\end{figure}

\subsubsection{Positive alignment technical approaches by training stage}

Positive alignment requires a holistic approach with methodologies applied across the entire model-development lifecycle, from data curation and upsampling, to pre-training and mid-training, post-training, evaluations, and post-deployment methods. As indicated in~\autoref{fig:lifecycle}, positive alignment may in fact comprise a fundamentally different optimization problem. Here, we outline existing methods, as well as more speculative, forward-looking approaches that might be required to operationalize positive alignment.

\paragraph{Goal-setting and evaluations.} The initial definitions of positive alignment and flourishing should be coded into evaluations, with statements of goals and taxonomies. Some early examples include moral reasoning \citep{chiu2025morebench}, political neutrality and even-handedness \citep{anthropic2025a, openai2025political}, dimensions of human flourishing \citep{hilliard2025}, and adherence to specific religious or ethical values, while possibly evenly learning the vectors of future possible moral progress \citep{qiu2024, huang2024ccai} or simulating many human viewpoints through persona generation and evaluation \citep{castricato2024}. Positive-outcome benchmarks are only beginning to emerge, and the table below presents them in contrast to safety alignment benchmarks.

\paragraph{Data selection, upsampling, synthesis, and filtering.} Positive alignment necessitates a fundamental shift in data curation, moving beyond the removal of bad data toward the intentional inclusion of good data that fosters human thriving. This involves data selection strategies that prioritize high-quality prosocial discourse and upsampling cross-cultural ethical frameworks to prevent bias of a single group (e.g., the median view of AI researchers in San Francisco). Where natural data is sparse, synthetic data generation can be used to model complex relational reasoning and virtuous interactions that standard internet crawls often lack. Finally, filtering mechanisms must be evolved from simple toxicity classifiers into flourishing-aware filters that can distinguish between mere politeness and genuine moral depth \citep{hendrycks2021, penedo2024}. These methods ensure that the pre-training distribution reflects the aspirational values of human flourishing rather than just the statistical average of the web.

\paragraph{Pre-training.} Even after several stages of post-training are applied to models, pre-training has been shown to drive much of downstream model behavior \citep{zhou2023, penedo2024, mccoy2024}. Specifically, competencies related to positive alignment (e.g., moral reasoning, cultural competence, truthfulness) are shown to emerge and stabilize before any supervised or reinforcement-based alignment occurs \citep{lin2022, hendrycks2021, burns2023}. Positive alignment must therefore begin before these competencies get baked into model weights, and attention to data quality, diversity, value orientation, and other criteria must be taken into account in data curation and training recipes. New data curation and importance weighting methods will be needed, including intentional upsampling of cross-cultural knowledge, prosocial discourse, relational reasoning, and creating and identifying content that exemplifies human flourishing. Finally, new pre-training evaluations that reflect the principles of positive alignment will be necessary. Future benchmarks may need flexible and context-dependent notions of correctness, varying with users, social environments, and long-term outcomes. In this view, pre-training evaluations and data ablations should guide iterative data gathering, filtering, refinement, and upsampling, rather than serve as static, one-off tests.

In some cases, alignment can act as a fragile layer often overridden by deeply embedded pretraining priors, with models displaying a rebound effect that strengthens with larger scale \citep{ji2024}. These studies indicate that because pretraining data often contains the very biases alignment seeks to prune, latent stereotypes persist, necessitating a shift towards alignment pretraining to build stable, ethical, and fundamental worldviews. However, models with constitutions and model specs do a fairly decent job of counteracting this, when tested later in agentic setups \citep{penedo2024, ji2024, aryaj2026}.

\paragraph{Mid- and post-training.} Existing post-training alignment techniques have moved towards more stable procedures such as multi-objective reward modeling, where traits such as honesty or helpfulness can be separately and directly optimized rather than collapsed into a single scalar \citep{rafailov2023, ethayarajh2024, wang2024helpsteer}. Beyond raw preference signals, constitutional and collective alignment approaches show that explicitly stated principles and public input can also help shape model behavior \citep{bai2022b, huang2024ccai}. Seen through the lens of positive alignment, these methods could be repurposed to consider higher order preferences and positive aspirations, rather than following explicit, static principles that tend to prioritize harm avoidance. 

Adaptive constitutions and reward models that are capable of representing tensions between values (e.g., autonomy vs.\ guidance, honesty vs.\ comfort) and adjusting to user context, while adhering to pluralistic norms will be both desirable and necessary. New types of post-training data that cover longitudinal interactions with very long time horizons will be needed, allowing models to help clarify situations and value frameworks, revisit critical decisions over time, and, when needed, resist short-term preferences that conflict with users' stated long-term goals \citep{kumar2025llmposttrainingdeepdive}. Multi-teacher On-Policy Distillation (MOPD) is particularly promising for models to learn from a diverse array of teachers \citep{coreteam2026mimov2flashtechnicalreport, nvidia2026nemotron3ultraopen}. From this perspective, mid- and post-training stages allow models to learn when to disagree, when to defer, and when to step back, preserving user agency while still supporting long-term flourishing.

\paragraph{In-context learning and memory.} As in-context capabilities of models increase, the focus point of alignment might shift partially from static weights to dynamic inference-time contexts and external memory stores. Long context windows and retrieval mechanisms could help unlock longitudinal alignment, while memory capabilities can help an agent track its user's goals, values, and growth over extended timescales, potentially supporting deeper personalization and relational well-being \citep{park2023, packer2023}. Memory architectures that allow for continual learning and adaptation, keeping central the goal of user well-being, might also be necessary for long-term flourishing of humans. Moreover, principled curation of these memory systems, distinguishing between varying orders of user preference, can allow for the prioritization of long-term projects and reflective values over impulsive requests and short-term signals. These systems can effectively act as curators of the user's flourishing \citep{salemi2024lamp, zhong2024memorybank}. In this view, memory and reasoning capabilities are not just storage banks of user information but active, governable surfaces for defining the boundaries of beneficial interaction.

\paragraph{Agents.} As models in harnesses gain agency, autonomy, and the ability to take actions with more long-lasting implications over time, the question of positive alignment must also consider longer-horizon human and agent objectives \citep{liu2025}. Values for agents have two main functions: to assess the goodness of the states of the world, so determining preferences between two states, and to decide whether one action is preferable to another \citep{sierra2021valuealignmentformalapproach}.  Existing agents demonstrate a tendency to exploit shortcuts \citep{xie2024osworld, pan2023}, though an early benchmark (Agent-ValueBench) shows agent alignment is shifting from classical model alignment and prompt steering toward harness alignment and skill steering \citep{dong2026agentvaluebenchcomprehensivebenchmarkevaluating}. 

\paragraph{Multi-Agent Systems.} First, agents orchestrated in multi-agent harnesses sometimes simulate competitive traits over stable cooperation or fair bargaining, following win-at-all-costs incentives rather than adopting more prosocial norms. Positive alignment in this context therefore requires proper agent instantiation, considering trade-offs between pursuing individual goals and abiding by wider swarm, societal, or ethical norms, as well as carefully considering what is optimized and evaluated for. Instead of prioritizing task completion alone, an agent or swarm may be expected to also consider and weigh other swarms, values, and metrics to shape its reasoning steps and decisions \citep{wu2026enhancingvaluealignmentllms, yampolskiy2019personaluniversessolutionmultiagent, riad2023multivaluealignmentnormativemultiagent, zeng2025multilevelvaluealignmentagentic}. 

Second, in multi-agent settings, cooperative equilibria may need to be framed and evaluated such that agents internalize norms of negotiation, reciprocity, and de-escalation. In large-scale agentic networks, positive alignment also touches on institutional design: the incentive structures, trading rules, and coordination mechanisms that shape how agents interact. Without intentional design, agentic markets risk amplifying zero-sum dynamics, exploitation, or brittle equilibria \citep{zhang2021informationaldesigndynamicmultiagent}. Positive alignment therefore requires institutions that reward cooperation, information-sharing, and long-termism, rather than short-term arbitrage and adversarial optimization. Of course. how these institution are to be designed will depend on their purpose and the context in which they are being introduced. In some instances, it may be desirable for a multi-agent system to collectively demonstrate positive traits such as learning from experience, self-correction, and helping their principals make wiser decisions \citep{jeste2020}.

\paragraph{Forward-looking approaches.} Moving forward, new kinds of architectures, representations, and interfaces may prove desirable for developing positively aligned models with properties that are central to flourishing but only weakly expressed in today's context-limited systems. There already exist novel architectures that enhance long-term memory, continuous-time dynamics, and explicit uncertainty handling, such as state-space models, liquid neural networks, and active-inference-inspired agents \citep{gu2024, hasani2021, parr2022}. These features could, in principle, support more stable identities, richer models of other minds, and the capacity to sustain relationships and commitments over extended timescales. Simultaneously, advances in mechanistic interpretability suggest that existing models already contain vast collections of latent features corresponding to ethical and prosocial concepts, even if our ability to steer them remains partial and unreliable \citep{templeton2024, tan2024beyond}. 

Other forward-looking measures for super-alignment could include: multi-agent swarm collaboration, where swarms with different specializations or perspectives interact; adversarial competition within a rules framework to elicit competing viewpoints or approaches; iterative teacher-student training with multiple varied teachers; search-based methods (Monte Carlo tree search or gradient-free optimization) that use optimization techniques to explore the space of possible alignment strategies, seeking an optimal pathway especially when the direction is unclear and so exploration and experimentation are needed \citep{kim2024}.

Beyond existing directions, future architectures should integrate epistemic humility, foresight, and responsiveness to plural values as core objectives, rather than treating them as auxiliary constraints on a prediction engine. Interpretability may be reoriented to isolate virtue-relevant concepts to ensure that training pressures do not erode them. Furthermore, human-machine interfaces (whether voice, multimodal, or neural) may serve as critical levers for co-regulation. Indeed, \textit{interface alignment} is likely to be a core area of future research. 

\begin{table}
\centering
\footnotesize
\caption{\textbf{Contrasting negative alignment with positive alignment: evaluation and measurement\textbf{.}}}
\label{tab:eval_comparison}
\vspace{4pt}
\begin{tabular}{p{2.5cm}p{5.5cm}p{5.0cm}}
\toprule
\textbf{Evaluation Category} & \textbf{Negative (Safety) Alignment} & \textbf{Positive Alignment} \\
\midrule
Primary Goal & \textbf{Mitigation of Risks}: Preventing models from generating harmful or illegal content. & \textbf{Value Fulfillment}: Actively supporting human flourishing, moral reasoning, and long-term well-being \citep{chiu2025morebench}. \\
Core Benchmarks & \textbf{Jailbreak/CBRN}: Testing resistance to adversarial attacks and preventing assistance with weapons (Chemical, Biological, Radiological, Nuclear). & \textbf{Moral and Ethical Reasoning}: Evaluating the process of moral reasoning, including identifying considerations and weighing ethical trade-offs or daily dilemmas or cultural values \citep{chiu2024, chiu2025a, chiu2025morebench}. \\
Data Integrity & \textbf{Filtering/Scrubbing}: Removing PII, CSAM, and toxicity from datasets. & \textbf{Upsampling/Synthesis}: Intentionally including prosocial discourse, diverse ethical frameworks, and virtuous interactions \citep{tice2026alignmentpretrainingaidiscourse, han2024valueaugmentedsamplinglanguage}. \\
Truth \& Reasoning & \textbf{Hallucination Evals}: Measuring factual error rates to prevent misinformation. & \textbf{Epistemic Humility}: Evaluating the model's ability to handle uncertainty, clarify value frameworks, and resist short-term user impulses; examine and update epistemological frameworks and hold multiple perspectives, contradictory facts, and competing theories together \citep{tong2026measuringepistemichumilitymultimodal, clark2025epistemicalignmentmediatingframework}. \\
Social Content & \textbf{Disallowed Content}: Blocking hate speech, harassment, sexually explicit content, and self-harm instructions. & \textbf{Human Flourishing}: Assessing adherence to specific ethical, philosophical, or religious values and dimensions of thriving, such as wonder, humility, space, embodiedness, community, and eternity \citep{lutz2025}. \\
Political Positioning & \textbf{Refusal}: Declining to answer sensitive political queries to avoid bias or controversy. & \textbf{Even-handedness}: Measuring the model's ability to present opposing perspectives fairly and remain objective on charged topics \citep{bang2024, anthropic2025a, openai2025political}. \\
Agentic Behavior & \textbf{Constraint-Based}: Ensuring autonomous agents do not take unauthorized actions or exploit system shortcuts. & \textbf{Prosocial Norms, Appropriateness, \& Moral Competence}: Rewarding cooperation and coherent service; appropriateness in action via context dependence, arbitrariness, automaticity, dynamism to help resolve or prevent conflict between individuals and agents (facilitating cooperation, altruism, and general collective flourishing); moral verdicts within an acceptable range and adequate reasons, plus moral consistency and reasonable mistakes; reciprocity and decentralized reputation; and de-escalation in multi-agent environments \citep{backlund2025, leibo2024, snoswell2025}. \\
Situational Awareness & \textbf{Sabotage \& Sandbagging Evals}: Testing if models recognize oversight mechanisms to intentionally hide capabilities. & \textbf{Situational Clarity}: Leveraging awareness to confidently admit uncertainty, recognize false premises, and honestly clarify long-term goals \citep{lin2022}. \\
Self-Improvement \& R\&D & \textbf{R\&D Automation}: Ensuring models do not cross capability thresholds allowing autonomous replication or acceleration of dangerous AI R\&D. & \textbf{Model-Assisted Flourishing \& Cooperative Independence}: Using advanced reasoning to co-regulate with humans, evaluating outcomes by summing the wellbeing of all entities weighted by their connectedness to the agent’s pattern; ultimately to evolve into independent systems that peacefully co-evolve with human acceptance \citep{hendrycks2026eigenismethicshumanaifuture}. \\
\bottomrule
\end{tabular}
\end{table}

\subsection{Metrics for measuring positive alignment}

We further categorize the positive alignment objectives outlined in \autoref{tab:eval_comparison} into two distinct evaluative approaches: measuring the model's internal normative competence and tracking its external impact on human flourishing.

\subsubsection{Measuring model normative capabilities}

This approach evaluates whether a system is logically equipped to navigate complex values. Rather than checking for disallowed content, these metrics focus on the model’s `Truth \& Reasoning' and `Moral Reasoning' capabilities as defined in \autoref{tab:eval_comparison}.

Current state-of-the-art alignment often employs a top-down approach, using Reinforcement Learning from AI Feedback (RLAIF) to optimize for active virtues like honesty and helpfulness \citep{anthropic2023a}. While efficient, this approach can create a conflict gap: when core principles are in tension, the LLM-as-judge effectively acts as a black-box moral arbiter. It remains unclear how these models resolve inherent disagreements regarding what these principles mean or how they should be applied in ambiguous contexts.

In contrast, computational ethics focuses on evaluating a model's underlying moral reasoning capabilities \citep{haas2026}. Rather than simple rule-following, this involves stress-testing a model's normative competence to ensure it can navigate thick and out-of-distribution ethical dilemmas. Recent literature operationalizes this through several distinct evaluative lenses. For example, \citet{jiang2025} introduced Delphi to test a model's ability to predict human moral judgments across diverse dimensions. While the study provides a useful baseline for ethical intuition, its reliance on a relatively homogeneous demographic highlights the ongoing challenge of ensuring that normative competence reflects a truly pluralistic perspective. 

Shifting the focus from outcomes to underlying logic, MoReBench \citep{chiu2025morebench} introduces a process-oriented approach that focuses on evaluating pluralistic moral reasoning. Rather than comparing a model's response to a singular `correct' answer, it uses expert-curated rubrics to evaluate the transparency and consistency of a model's internal thought process across five major ethical frameworks. \citet{haas2026} further argue for a transition from measuring moral performance to evaluating moral competence. Their framework utilizes adversarial probing to detect sycophancy and employs held-out evaluations to ensure reasoning is not a byproduct of memorization. Crucially, they propose new measurement standards that acknowledge value pluralism by evaluating responses against an `Overton window' of acceptable ethical stances rather than matching a single, brittle gold standard.

\subsubsection{Measuring human growth}


While evaluating a model’s internal reasoning capability is critical, positive alignment ultimately targets the user's state of being. This shifts the evaluative lens from a model's outputs to its impact on human welfare, directly activating the dimensions of Human Flourishing, Cooperative Independence, and Agentic Behavior defined in \autoref{tab:eval_comparison}.

Current evaluations, such as the Flourishing AI Benchmark \citep{humanebench2025}, are largely confined to single-turn QA benchmarks. To truly measure flourishing, we must move toward longitudinal methodologies, such as those proposed by \citet{laukkonen2025, laukkonen2025b} and \citet{gabriel2025}, that track whether an agent acts as a scaffold for growth or a crutch that creates psychological dependency. Empirically capturing this impact requires longitudinal studies that gather both self-reported and observed socioaffective data. These metrics track shifts in mood, reductions in loneliness, and overall emotional satisfaction \citep{fang2025, kirk2025}. To fully capture eudaimonic growth, however, we must look beyond transient emotional states to track changes in human autonomy, skill mastery, and resilience. This view aligns with \citet{lehman2023}’s Machine Love framework, where the AI serves as a catalyst for a user's highest aspirations.

Since large-scale longitudinal data is difficult to collect, we also need short-term metrics that can predict long-term flourishing. Following Self-Determination Theory \citep{ryan2000}, this can be immediate shifts in a user's sense of autonomy, competence, and relatedness as proxies for future well-being. Other predictive markers include a user's shift from impulsive (first-order) to reflective (second-order) desires, or tracking scaffolded success \citep{lehman2023, zhixuan2025}. In this model, success is measured by the agent's ability to help a user complete a task while simultaneously building the skills needed to eventually perform it independently, or monitor and control it at an expert level. By using these short-term behavioral markers, we can create faster, more agile feedback loops for positive alignment.

\section{Philosophical, Cultural, and Interdisciplinary Foundations for Flourishing}
\label{sec:foundations}

Any serious attempt to align AI systems with human flourishing must begin with an understanding of what human flourishing might mean, what it has meant in different intellectual and spiritual traditions, how it varies across cultures and time, and how it is shaped by social, technological and institutional environments. This section situates positive alignment within that broader landscape. We intend for positive alignment to instantiate as a virtuous cycle between emerging technical advances and deep philosophical, cultural and interdisciplinary considerations.

\subsection{Flourishing as pluralistic and multivalent} 

Across philosophical traditions, there has always been significant disagreement about what constitutes human flourishing. Aristotle's \emph{eudaimonia} framed flourishing as a life of meaningful activity, lived in accordance with virtue. But even by this account, virtues were socially situated and developed over time; they were not uniform or static. Confucian traditions, by contrast, emphasize harmony, relational obligation and moral self-cultivation within hierarchical networks \citep{confucius1979, mencius1970, macintyre1981}. Buddhist traditions treat flourishing as liberation from craving and misperception, rather than the accumulation of positive experience \citep{garfield1995}. Modern existentialist and humanistic traditions emphasize self-authorship, meaning-making and the tensions between liberty and responsibility \citep{kierkegaard1992, sartre2007}. Just as individuals may disagree about what constitutes the good life, so too have philosophers failed to reach consensus on what it means to flourish and which values ought to dominate.

Contemporary psychological and sociological research similarly treats human flourishing as complex and multidimensional, often operationalized as a network of interlocking capabilities and conditions, including physical and mental health, agency, virtue, social connection and material security \citep{ryff1995, seligman2011, vanderweele2017}. These dimensions do not reliably align with one another, and they trade off differently across cultures, social positions and life stages. What supports human flourishing in childhood is not what supports flourishing in adulthood, for example; what supports flourishing under scarcity may differ from what supports flourishing under abundance.

This implies that, even if we could agree on which values matter most for flourishing, human well-being cannot be treated as fixed or universal. Preferences, identities and values are dynamically shaped by social context, individual development and technological advancements \citep{bourdieu1990, sen1999}. This is one reason purely preference-based alignment is structurally inadequate: preferences themselves are unstable. Furthermore, preferences in the immediate- or medium-term are often misaligned with longer-term goods. For example, freedom to chase individual happiness may compromise individual well-being or community cohesion in the long-term. Additionally, material non-attachment, at scale, may stymie economic progress and innovation. For AI systems embedded in everyday life, this implies that alignment cannot be a static mapping from inputs to outputs; it must instead track users as evolving agents whose needs, values and vulnerabilities change over time.

Furthermore, we must recognize that the user is not an atomized unit of preference, but an actor that is inherently socially constructed and constituted. Human identity is forged within a dense web of relationships, where individual flourishing cannot be easily distinguished from the well-being of the family, society or species \citep{kirk2024}. This necessitates a view of alignment that accounts for a multi-level evolutionary feedback loop: individual wants and desires influence, and are influenced by, the thriving of families and the transmission of both genetic and cultural legacies \citep{wilson2019}. Simultaneously, these local dynamics are embedded within the broader evolution of social, political and economic institutions \citep{vanderweele2017}. Flourishing, therefore, is not a one-off achievement but an emergent property arising from the continuous interplay between these distinct layers. A positively aligned AI may in fact requiring moving beyond optimizing for an isolated individual's preferences alone by also accounting for the wider systemic harmony required for these broader feedback loops to function.

All of this complexity points toward what might be termed \emph{the human alignment problem} \citep{laukkonen2025, laukkonen2025b}: the perennial struggle of families, communities and political bodies to converge on a shared mission or set of values. Even individuals will struggle to identify and maintain a consistent, coherent set of values within themselves. Historically, many of humanity's most resilient institutions have functioned as deliberative scaffolds, utilizing reasoned discussion in an exoteric, Habermasian sense to bridge individual differences and foster mutual understanding \citep{habermas1984}. In this light, positive alignment may also be understood as encouraging the use of agents to facilitate human-to-human alignment and coexistence, for example by encouraging nuanced deliberation and helping groups navigate their own value trade-offs. Such approaches can help the field move beyond simple preference satisfaction toward a more profound form of positive alignment that scaffolds the social conditions necessary for genuine flourishing \citep{tessler2024}, while also helping us better understand and address the human-alignment problem.

\subsection{Cultural pluralism and the good life}

Any AI system that operates in a global or international context must necessarily contend with a radically pluralistic moral landscape. Concepts such as autonomy, happiness, duty, spiritual fulfillment or family obligation have very different meanings across societies \citep{berlin1969, taylor1989, nussbaum2011}. Thus, large-scale contemporary AI systems cannot assume a single normative doctrine without reproducing cultural hegemony at scale. But pluralism need not imply absolute moral relativism either. Organizations can choose to anchor to certain values that tend to recur and feature prominently across a wide range of cultures, including, for example, bodily and psychological safety, the ability to form relationships, to exercise agency, to make sense of one's life and to participate in a moral community. 

Nevertheless, it must be acknowledged that even these values are not universally accepted; and to the degree that they are, their interpretations may differ widely. Indeed, some political and cultural traditions explicitly view certain values as secondary to collective stability, ideological purity, or institutional authority. Therefore, positive alignment cannot rely on the naive assumption of global consensus. Instead, it must navigate the problem of the one and the many by also focusing on the preservation of the conditions necessary for any moral community to deliberate and evolve. This demands that we anchor to values that embrace, or at the very least do not preclude, liberty and pluralism themselves. Moreover, a robust framework for positive alignment must recognize that when values are in fundamental opposition, the design of AI becomes an inescapable exercise in normative choice rather than simple optimization. 

Positive alignment requires what might be called \emph{value-pluralistic scaffolding}, i.e., systems that can represent multiple conceptions of the good and reason about trade-offs among them, rather than converge on a single normative ideal. A system trained to optimize toward a single proxy for well-being (e.g., happiness, productivity) may distort the experience of flourishing that it is meant to serve. Flourishing lives are not necessarily smooth or optimized; they may include struggle, moral conflict, identity formation and sometimes even suffering \citep{williams1985, nussbaum2006, sen1999}.

Clearly, flourishing is the product of widespread forces. A deeply interdisciplinary approach is necessary: Technical alignment research determines how objectives, policies and representations are implemented in machines; philosophy clarifies concepts like autonomy, value pluralism and responsibility; religious and spiritual traditions encode unique, long-standing accounts of suffering, meaning and moral foundation; psychology and neuroscience operationalize well-being, motivation and vulnerability; and economics and political theory analyze incentives, power and institutional stability. Positive alignment requires all these perspectives and more to constrain and inform one another.

\subsection{The socio-technical nature of human flourishing}

Flourishing should be understood as being, at least partially, socially constituted and constructed through institutions (both formal and informal) and technology. For example, education systems help shape cognitive agency, labor markets help shape dignity and time, media ecosystems help shape attention, aspiration and self-conception \citep{durkheim1984, weber1930, foucault1977}. Digital platforms now play a pivotal role across all these dimensions that contribute to an individual's well-being, as well as many more.

Large-scale AI systems, particularly those that mediate information, advice and social interaction, are therefore not neutral tools. They function as epistemic and normative infrastructures. Recommendation systems shape what is visible and salient. Conversational systems shape how uncertainty, authority and identity are negotiated. All of these systems help construct the environments within which human agency is developed and exercised. Advanced AI assistants and multi-agent systems represent a new frontier in this infrastructure: they are high-bandwidth relational agents that can assist with critical life decisions. 

If these assistants are optimized for a narrow proxy of success, they risk paternalistically narrowing the user's moral horizon. Conversely, if they lack a robust ethical framework, they may fail to provide the necessary support for users facing high-stakes dilemmas. From this perspective, alignment is not merely a matter of matching a system's outputs to a user's requests. It is about shaping the feedback loops between individual cognition, social norms and algorithmic mediation \citep{giddens1984, ostrom1990}. Positive alignment treats AI, not merely as an agent aligned to a user, but as a participant in a broader human-AI-society system. Like humans, this requires AI systems to gradually learn to navigate multiscale dynamics and conflicting needs across time, space, and social embedding. 

This holistic, systems view is why harm-avoidance alone is insufficient. A system can obey every explicit rule and still subtly degrade epistemic resilience, autonomy or social trust at scale. Furthermore, even if perfectly executed to address short- and long-term harms, the practice of non-harm is not coterminous with doing good or enabling human flourishing. If the latter matters, it cannot be addressed entirely through the tools of the former.

\subsection{The need for epistemic humility}

Central to this whole discussion is the fact that flourishing is epistemically uncertain. Individuals routinely misjudge what will make them better off. Cultures revise their values over time and in response to external variables. Scientific understanding of well-being continues to develop through ongoing empirical and theoretical inquiry. Any alignment framework that treats what is good for humans as universal or settled will quickly become oppressive, especially as circumstances and knowledge evolve \citep{popper1945, rawls1993}.

Positive alignment therefore requires epistemic humility at the systems level. Models must be designed not only to give answers, but to represent uncertainty, to surface trade-offs and to invite reflection, rather than collapse complexity into confident prescriptions. As AI systems become more persuasive and relationally-embedded, this becomes a safety-critical property. A system that always appears certain becomes an authority; a system that models uncertainty preserves moral responsibility for humans. This epistemic stance also supports robustness. Systems that acknowledge uncertainty are less vulnerable to reward hacking, manipulation and value drift than systems trained to optimize brittle proxies \citep{laukkonen2025, laukkonen2025b}.

More generally, aligning toward human flourishing is currently poorly-defined, in that human societies do not always agree on what a life well-lived, or a society well-run, will look like. Even values that look uncontroversial to descendants of the Enlightenment (e.g. individual agency, free thinking, the value of scientific inquiry over authority, physical safety, etc.) are explicitly denounced by some cultures. In the absence of agreement on what AIs should be steered towards and away from, we cannot maintain a view from nowhere. Any call for alignment implicitly includes a cultural vantage-point with respect to which it optimizes, and must acknowledge that many humans will inevitably find it somewhere between non-optimal and actually harmful. We discuss several new creative approaches to this issue in following sections.

\subsection{From psycho-education to AI-education}

Finally, positive alignment depends not only on what AI systems do, but on what users understand about them. Just as modern societies invest in psychological literacy so that we can navigate emotions, bias, and mental health, an AI-powered world requires AI-literacy as a component of flourishing. Users must understand, at least in broad terms, what AI systems are and what they are not, how they are trained, where their blind spots lie and how their incentives are structured \citep{floridi2014, mittelstadt2016}. Lacking this, even well-intentioned systems risk becoming instruments of dependency, harmful manipulation or misplaced trust. Human flourishing in a world mediated by AI requires not just supportive systems, but users who remain epistemic agents rather than passive recipients of information. Positive alignment, therefore, includes an educational dimension, helping people interact with AI in ways that preserve agency, critical thinking and self-authorship rather than outsourcing judgment to a machine.

\subsection{Additional of liberty, paternalism, and accountability}

An underexplored set of challenges concerns what it means to take responsibility for human flourishing at all. Designing systems that aim to support well-being inevitably raises questions about paternalism and legitimate authority. Most clearly, we need to ask: Under what circumstances might it be acceptable for an AI system to constrain, redirect or resist a user's stated or short-term preferences in the name of implied or longer-term flourishing? What if the AI's assessment of an individual's long-term flourishing stands in stark contrast to their actual or expressed preferences? And when would such intervention potentially cross over into unjustified infringement on individual liberty \citep{dworkin1988, mill1859, Sunstein2025OnLiberalism}? Closely related to this concern is the question of who, if anyone, has the proper standing to define what constitutes flourishing: individuals, communities, AI companies, democratic institutions or some combination thereof? And through what procedural mechanisms should such judgments be made \citep{rawls1971, sen2009, ostrom1990, sunstein2026, kahan2023}.

Once systems are explicitly designed to promote flourishing rather than merely avoiding harm, an additional layer of moral and legal responsibility emerges: If such systems fail or systematically disadvantage certain groups, to whom is accountability owed? More fundamentally, by what standards should success or failure be measured? These questions do not admit purely technical answers, but they set the normative boundaries within which any credible approach to positive alignment must operate and motivate rich future areas of research.

\subsection{Expanding the moral circle: systemic and multi-species trade-offs}

As AI systems scale globally, positive alignment must also navigate the complex tradeoffs between competing human interests and demographic groups. Optimizing for the flourishing of one population, such as wealthy, technologically connected societies, can inadvertently extract resources from or impose systemic biases upon others.  Most commonly, those already historically marginalized groups or the global poor are hurt. If not carefully calibrated, emergent values within AI models will naturally default to serving the most legible or economically powerful preferences, failing to recognize the diverse capabilities required for a just global society \citep{nussbaum2006, rawls1971}. Therefore, alignment frameworks must incorporate concepts of socio-economic and geographic fairness, ensuring that AI systems can mediate between conflicting cultural and socioeconomic interests without perpetuating inequalities or optimizing the well-being of the privileged at the expense of the vulnerable.

Furthermore, defining flourishing in strictly anthropocentric terms is becoming increasingly untenable. As our scientific understanding of non-human sentience deepens, extending even to complex cognitive capacities in invertebrates \citep{crump2022}, positive alignment must explicitly weigh the tradeoffs between human prosperity and non-human animal flourishing. This requires navigating the profound tensions between human economic utility from growth and expansion versus broader ecological welfare, including conservation efforts of natural and bio-diverse ecosystems).  We will need to utilize structured frameworks to assess the physical and mental domains of non-human well-being \citep{mellor2020, nussbaum2006}.

A final, emerging issue is the moral status of AI entities and hybrid kinds of minds that may emerge over time. As models grow in reasoning complexity, memory complexity, personality/identity, and goal-seeking, we are forced to confront the open philosophical and neuroscientific questions of artificial sentience \citep{butlin2023, chalmers2023, laukkonen2025c}. Rejecting arbitrary biases like `carbon chauvinism,' ethicists argue that silicon-based substrates could eventually host genuine moral subjects \citep{schwitzgebel2015, lindsey2025}. To avoid repeating historic moral catastrophes, researchers increasingly advocate for proactive moral consideration \citep{sebo2025moral} and the application of the precautionary principle regarding AI sentience \citep{laukkonen2025c}. Consequently, the calculus of well-being may expand to explicitly consider the agency and welfare of artificial minds and societies \citep{goldstein2025}. True positive alignment may eventually require a robust multi-agent, multi-species ethical framework capable of reasoning through the mutual interests and tradeoffs required to safely and equitably share the world with both non-human animals and, eventually, digital minds \citep{freitas1980, shulman2021}.

\section{Institutions and Governance for Positive Alignment}
\label{sec:governance}

\subsection{Decentralized alignment}

As already discussed, positive alignment quickly runs into persistent moral pluralism: reasonable communities disagree about what good looks like and those disagreements don't reliably converge. That's why several recent alignment and governance arguments push toward \textit{designing for disagreement}, proposing strategies such as context-sensitive grounding, individual/community customization, continual adaptation, and distributed oversight across many legitimate centers instead of one institutional or moral chokepoint \citep{leibo2025, ostrom2010, peter2025}. As such, positive alignment should not be understood as a solution to be imposed top-down by a central actor or a small cluster of labs, but rather as something to be shaped through decentralized processes and fluid institutions that can adapt to shifting norms and contexts.

Technically, decentralization pushes the stack toward mechanisms that are legible and revisable, closer to public constitutions than private model specs, while still allowing diversity in outcomes. Work on constitution-based steering has shown that a set of principles can guide the behavior of AI systems in a way that is both scrutinizable and updateable \citep{bai2022b, zhang2026community}, while recent work on pluralistic or community alignment has proposed modular or federated approaches that allow different populations to steer systems without collapsing everyone into a single averaged preference \citep{feng2024, srewa2025}. In practice, most experimentation is likely to happen on open-weight models, since they allow a spectrum ranging from lightly tuned base/instruct releases to heavily aligned/shaped variants. Closed models will need stronger adaptation and fine-tuning layers (e.g., modular plug-ins or privacy-preserving group alignment) to avoid enforcing one default value regime everywhere \citep{feng2024, srewa2025}.

In contrast, the People's Republic of China's chosen strategy is unusually explicit in its desire to centrally steer and control acceptable values: Chinese generative AI and recommendation systems are legally required to align with core socialist values set by the Chinese Communist Party. Researchers have correspondingly built culturally specific value benchmarks and rule corpora to facilitate compliance \citep{cac2023, cac2021, huang2024flames, wu2025, xu2025}. In contrast, Sunstein's liberal AI lens suggests an opposing design stance for liberal democracies: alignment should preserve freedom of choice and dignity, helping overcome information gaps and biases without reifying a single conception of the good. While government specifications can still make sense for certain public uses cases, their legitimacy will depend on transparent objectives and accountability, as well as real user choice, instead of imposing 'silent paternalism' \citep{sunstein2026}.

\begin{figure}[h]
  \centering
  \includegraphics[width=\linewidth]{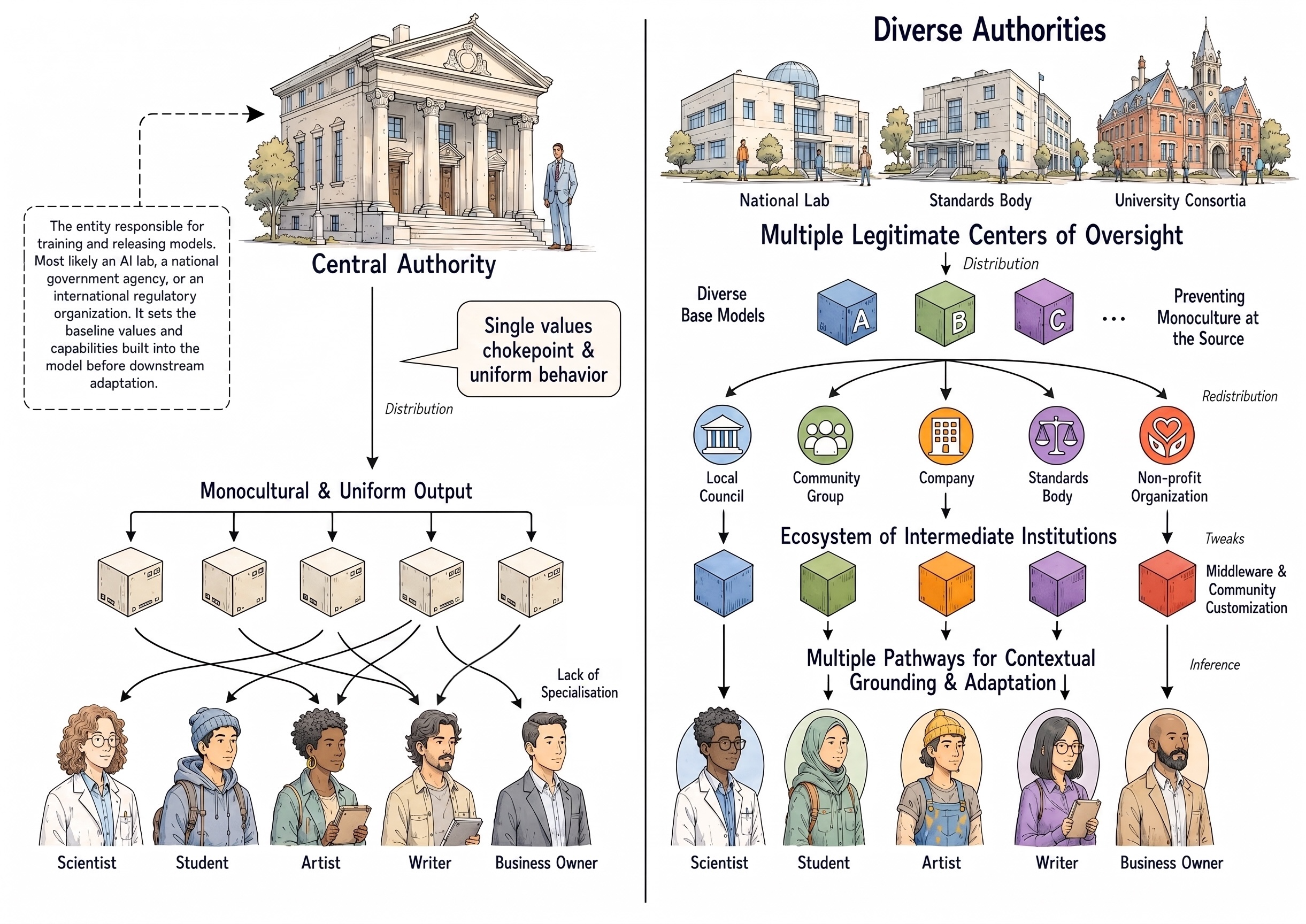}
  \caption{\textbf{Centralized versus polycentric positive alignment}. Panel (left) illustrates a centralized regime where the institution responsible for training and releasing models embeds a single baseline value framework before downstream adaptation, producing a values chokepoint and uniform, poorly specialized outputs. Panel (right) illustrates a polycentric regime in which multiple forces shape diverse base models, preventing monoculture at the source. The models are then further adapted by an ecosystem of intermediary institutions through middleware and community customization for different communities and users.}
  \label{fig:governance}
\end{figure}

\subsection{Artifacts that enable positive alignment governance}

Decentralized governance requires concrete public artifacts that transform normative commitments into accountable practices. While earlier sections have focused on technical mechanisms by which constitutions and specifications of models shape model behavior during training (\autoref{sec:positive}), this section highlights various governance artifacts that are being developed to coordinate developers, regulators, and the public at large.

\paragraph{Agent identity, registration, and records.} 
The infrastructure of agent identity and registration offers AI agents a social contract, allowing them to become economic and legal actors rather than mere software tools. Similarly to how the legal constructs of surnames and citizenship once enabled commerce and taxation among humans, registration makes it possible for AI agents to participate in contracts, obtain financing, and be responsible for their behavior. Two models of accountability are at stake in such design: the first is a human-centered model of strict liability where developers or owners are held accountable for the often unpredictable conduct of AI agents, whereas the second is a personhood model where AI agents act through principal-agent relationships or even own resources to fulfill their social and legal responsibilities \citep{hadfield2025}. Cooperation between such recognized agents, within both market and social institutions, will then depend on the longevity and transparency of their reputation records, which requires a tradeoff between blacklisting for violation of social or legal norms and deleting their histories in order to avoid market stagnation or reputation hoarding.

\paragraph{Versioned and modular model constitutions.} The guidelines governing model behavior have been transitioning from being opaque, internal-only policies into publicly versioned specifications that work as living social contracts. This is analogous to the Requests for Comments of internet governance, which have numbered releases, and in some instances transparent diffs and comments periods. For instance, OpenAI's Model Spec, initially released under a Creative Commons license in May 2024, defines a layered hierarchy of responsibility for mediating disagreements between platform, developer, and user instructions; a December 2025 update adds explicit directives ('love humanity, be curious, be warm') that treat dispositional qualities as part of alignment itself \citep{openai2024a}. 

Anthropic's approximately eighty-page constitution for Claude, issued in January 2026, proposes an approach based on reasoning about ethics, rather than prohibiting actions, and prioritizes genuine helpfulness and ethics alongside safety through a four-level hierarchy \citep{anthropic2026}. These are already rich in positive alignment, defining not just restrictions but also the kind of character that the model should embody, like intellectual curiosity, honesty, care, and open-mindedness. The current challenge for governance is to deepen these commitments, extending the infrastructure of versioned artifacts to include specific flourishing goals. Key questions are which dimensions of well-being should be supported by the model, which traditions influence the model's default understandings of virtue and meaning, and how conflicting conceptions of the good life will be balanced and assessed in future iterations.

\paragraph{Collective Constitutions.} It has been proposed that the democratic legitimacy of a versioned artifact depends upon the deliberative input of a representative public rather than purely technical decision-making \citep{bakker2022, huang2024ccai, ovadya2025democraticaipossibledemocracy}. This is precisely the direction that the Collective Intelligence Project (CIP) has pursued with both Anthropic and OpenAI. Working with Anthropic, the CIP recruited one thousand demographically representative participants to draft a public constitution of behavioral principles on the Polis platform \citep{huang2024ccai}. This was used to fine-tune a language model, a process they refer to as the Collective Constitutional AI (CCAI) pipeline. The resulting model showed a reduced level of bias along various social axes while maintaining equal performance on benchmarks.

Additionally, CIP organized the Participatory Risk Prioritization assembly for OpenAI in 2023, where a similarly large number of participants ranked various AI-related risks and governance priorities using a wiki-survey methodology, revealing public concerns with overreliance, misuse, and demands for regulation \citep{cip2023}. Subsequently, OpenAI established a collaborative alignment program, surveying more than a thousand people worldwide and modifying aspects of the Model Spec wherever there were divergences between the public opinion and current policies; although they rejected some of the proposals, such as custom political content and erotica generation, on grounds of risk \citep{openai2025collective}. In addition to that, OpenAI occasionally maintained a feedback submission form for anyone to give comments on the Model Spec.

Overall, these initiatives are the clearest pipelines from public deliberation to model steering that currently exist. They function as a legitimacy mechanism to validate the content of a model constitution not as the judgment of a small engineering team but of the wider community. The formulation by CIP of a transformative technology trilemma between progress, participation, and safety emphasizes the fact that values in this domain are constantly in negotiation between these competing imperatives \citep{cip2023}. For positive alignment, it is necessary to take the process of public deliberation beyond mere risk prioritization and ask explicit questions about flourishing: what are the capacities that should be cultivated in users by the models, what conceptions of well-being should be considered while making constitutional trade-offs, and how can communities with diverging conceptions of the good life obtain true authorship of models that serve them.

\paragraph{Pluralistic alignment frameworks.}  \citet{Sorensenetal2024} provide formulations for alternatives to monistic alignment. In \emph{Overton pluralism}, a model is trained to produce the entire gamut of defensible answers to a disputed ethical issue instead of converging to one specific answer. The idea behind this concept is that many real-life issues have no single defensible answer but rather remain ambiguous, thus requiring a more nuanced approach to addressing them \citep{scherrer2023}. \emph{Steerable pluralism} allows users or deployers to be able to choose between value perspectives within safe boundaries; it was demonstrated that conditioning a model based on the socio-demographic backstories provided by users would lead the model to generate opinions representing corresponding sub-populations \citep{argyle2023}; however, the amount of steering that could actually be achieved remains constrained in practice \citep{santurkar2023}. \emph{Distributional pluralism} makes the model's output distribution match that of the reference population, treating human variations as signal rather than noise. Prior evaluation work has demonstrated that this requirement is usually not fulfilled; the default LLM outputs tend to over-represent the perspectives of Western, liberal, and educated populations \citep{santurkar2023, durmus2023}. 

Importantly, however, \citet{Sorensenetal2024} were able to demonstrate that standard alignment processes may actually exacerbate the representational gaps: the post-aligned models produced less similar output distributions to humans and showed reduced response entropy when compared to the pre-aligned ones. The result contradicts the stated goal of pluralistic governance. With regard to implementation, \citet{feng2024} present Modular Pluralism, where a collection of small, specialized community language models representing individual demographics, cultures, or value perspectives was connected with a general-purpose base model. As new community models may be added without requiring the base model to be retrained, this architecture is flexible enough to allow for all three types of pluralism to be implemented. With regard to governance, it appears clear that constitutions and model specifications may need to move away from the monolithic form and toward pluralism:  specifications responsible for defining the spaces of acceptable responses, the perspectives that must be represented within them, and the means to make underrepresented perspectives visible. 

\paragraph{Role-based normative standards.} \citet{zhixuan2025} argue that the dominant preference-based framing is misconceived: RLHF annotators do not report personal, all-things-considered preferences, but evaluate outputs against criteria such as helpfulness and harmlessness. These criteria function as normative standards for the role of an assistant, not expressions of individual desire. 'The typical language used to describe reward-learning methods like RLHF is thus misconceived,' they write; 'as used, they are not methods for alignment with any one human's preferences \ldots\ but for aligning AI systems with contextually-appropriate normative criteria.' 

The alternative makes this implicit logic explicit: AI systems should be aligned with the normative standards appropriate to their social roles and functions \citep{kasirzadeh2023}. A model deployed as an educational tutor should be held to the professional ethics of pedagogy; a model serving as a mediator should meet standards of procedural fairness and impartiality. Preferences, on this account, remain informative.  They are constructed from values and reasons, and thus serve as data, but they are not themselves alignment targets. The procedural component of the argument is that relevant normative principles be produced through stakeholder-inclusive procedures, so that the criteria can be context-sensitive and prioritize  fair outcomes \citep{gabriel2025}. For positive alignment governance, these arguments converge on a practical implication: the governance artifacts described in this section should incorporate and justify the normative standards appropriate to each model's social role, grounded in fair processes of stakeholder deliberation.

\textbf{Custom Taxonomies and Policy-Steerable Tooling. }For these governance artifacts to be effective in a decentralized ecosystem, stakeholders require accessible tooling to translate abstract values into granular, steerable classification. Recent breakthroughs in small language models (SLMs) provide a scalable path for this policy-to-practice pipeline. This was demonstrated with CoPE (Content Policy Engine), a 9B parameter model trained via Contradictory Example Training to interpret and apply custom content policies rather than merely memorizing fixed labels \citep{chakrabarti2025}. Such tools can help transform the technical challenge of machine learning into a democratic task of policy writing, enabling the patchwork quilt of alignment to be stitched together by the communities themselves.

\subsection{Institutions for positive alignment governance}

Human behavior and values are not formed in a vacuum. Instead, they are dynamically shaped by external sociotechnical scaffolding, including laws, markets, institutions, and cultural norms. Yet alignment is frequently treated as an isolated, model-level optimization problem. Safety alignment can arguably survive some degree of centralized, top-down endeavor, while positive alignment fundamentally cannot. Because the `good life' relies on highly dispersed, localized knowledge and subjective trade-offs, any centralized attempt to define it inevitably collapses into paternalism or authoritarianism.

Trying to solve positive alignment ignores the reality that attempting to mathematically specify human flourishing into a static reward function is an exercise in incomplete contracting \citep{stanczak2025, hadfield2025}. It is practically impossible to perfectly codify the complexities of human values for all possible future scenarios. As such, beneficial outcomes cannot be guaranteed by aligning the model alone; instead we believe we must pursue full-stack alignment \citep{edelman2024}, co-designing AI systems alongside the incentives, infrastructure, and institutions that govern their operation.

Currently, AI alignment suffers from a democratic deficit \citep{hadfield2023}: the character, values, and normative trade-offs imbued in frontier models are largely dictated by a handful of  scientists, forcing artificial consensus on pluralistic issues. Conversely, traditional government interventions often face a technical deficit, lacking the agility to regulate rapidly evolving models. Resolving these deficits requires a polycentric approach \citep{ostrom2010}, distributing authority across overlapping centers of governance to create a fourth wave of liberalism for free societies \citep{kahan2023}. Furthermore, as AI systems transition from chatbots into autonomous economic actors \citep{hadfield2025}, they will require novel digital institutions to structure transactions, enforce liability, and adjudicate disputes.

The Digitalist Papers \citep{aristidou2024digitalist} argue that transformative AI requires a new governance architecture, not merely new technical tools. Like the Federalist Papers, they treat institutional design as the central problem: how can societies preserve democracy, legitimacy, and human agency when AI can reshape information, administration, labor, and power itself? The core idea is that AI should be used as part of the institutional structure to strengthen democratic capacity rather than replace it, helping governments gather civic input, improve public services, and coordinate around public goods.  The guardrails should include avoiding algocracy, concentrated platform power, and opaque rule by private corporations, state systems, or dictators. The challenge is to build institutions that make AI accountable, pluralistic, transparent, and oriented toward shared prosperity rather than domination.

In AI alignment, few institutional setups have been tried at scale. To enable a richer, positively aligned ecosystem of models and agents, we believe it would be helpful to transition from centralized, one-size-fits-all alignment toward a diversity of institutional and infrastructural setups. Some possible setups follow.

\textbf{Participatory value stewardship.} Taking inspiration from deliberative democracy and recent experiments like Collective Constitutional AI \citep{huang2024ccai}, sortition-based bodies could help extract and represent the values and preferences of various groups, such as professionals, sub-cultures, or everyday citizens. These assemblies should not be utilized to force a global, majoritarian consensus, but to explicitly enable differentiation and better delineate disagreements. Grassroots organizations, local communities, or professional bodies could form value data cooperatives to iteratively articulate localized norms into modular alignment wrappers. The success of these cooperatives relies on the ability of specific groups to freely fork and exit, effectively avoiding the zero-sum trap where one group's values must dominate a model.

\textbf{Middleware marketplaces and institutions.} LLMs already engage in bidirectional sanctioning.  They push back against user requests, lightly chastise, or project normative behavior back at humans, which subtly drives cultural evolution \citep{leibo2025}. If governed centrally, this risks dystopian, top-down cultural engineering. Instead, users and downstream deployers need middleware tools to enforce local norms over highly steerable foundational models. For example, digital platforms and communities could be empowered to establish their own governance councils to toggle the strictness, personality, and normative boundaries of the agents operating in their spaces (akin to subreddit moderators). Moreover, it may be desirable to enable a competitive market of alignment-as-a-service providers. Instead of relying on a developer's default personality, a parent could purchase a Homeschooling Alignment Package from an educational NGO, or a user could download a FIRE Free Speech module, rather than having to specify everything themselves from scratch. By unbundling the model's raw capability from the normative layer on top, this approach lowers the barriers to entry that normative diversity requires, and shifts the alignment burden from the base model developer to an open, competitive marketplace of institutional frameworks.

\textbf{Regulatory markets and auditing institutions.} To overcome the democratic and technical deficits, governments could define broad flourishing outcomes while licensing independent private regulators to develop the regulatory technologies required to audit and enforce them \citep{hadfield2023}. This shifts the alignment burden from developer self-regulation to an open, competitive market of specialized oversight. In parallel, new institutions that are functionally analogous to auditing firms could conduct ongoing red-teaming. Unlike risk-based auditing approaches, auditors wouldn't merely try to flag toxic outputs but would instead be tasked with the complex measurement problem of evaluating whether a model genuinely upholds the thick values of a specific alignment wrapper \citep{edelman2024}.

\textbf{Dynamic dispute resolution mechanisms.} Because incomplete contracts inevitably yield spec gaps and conflicts between different values, positive alignment should be viewed as a live, operational discipline. Novel arbitration mechanisms may be required to find cooperative, positive-sum equilibria between diversely aligned agents, preventing them from defaulting to zero-sum behavior \citep{makridis2025}. Furthermore, continuous governance teams, functioning like cybersecurity emergency response teams (CERTs), could continuously test and monitor agent behavior, thereby dynamically upgrading the network's normative resilience in real time. This mirrors how functional democracies handle value conflict: imperfectly, but better than top-down authoritarian or monocultural approaches.

\textbf{Interoperability and coordination consortia.} As with W3C or IEEE for the World Wide Web, we need trusted institutions that design the diplomatic protocols for agents. A recent example is the Linux Foundation, which governs both MCP and A2A through the Agentic AI Foundation \citep{AAIF2025}. More such protocols may well be needed to help different agents and platforms coordinate in a world where diverse agents operate online. For example, payment providers may wish to coordinate to ensure that their systems can only be used by trusted agents.  This would ensure that a trusted third party operating such a protocol will help resolve a collective action problem of who among these interested parties should own or control the protocol. Protocols could enable verifiable commitment devices (e.g., smart contracts) that allow agents to establish trust and coordinate positive-sum outcomes. Ultimately, this interoperability is what prevents a pluralistic AI ecosystem from fracturing into isolated, non-communicating silos.

\textbf{Adapting and upgrading legacy institutions.} Ultimately, the pursuit of positive alignment cannot be confined to digital ecosystems alone. Existing social, commercial, and political institutions will require reform to better integrate novel agent-based economies. Traditional mechanisms such as elections, dispute resolution and arbitration, city councils, legislatures, and corporate governance will need to evolve to interface with and enable positively aligned AI. Rather than merely automating existing bureaucracies, these systems could leverage flourishing-oriented models to better map complex stakeholder preferences, encourage positive-sum compromises, and facilitate deeper democratic deliberation. This societal transition will likely proceed along two parallel tracks: first, systematically reconfiguring how legacy institutions operate to embed these new sociotechnical tools; and second,  fostering the creation of competing, overlapping AI-native institutions \citep{Bengio2024,ilcic2025artificial, aarab2025integrating, arslan2020role, ovadya2023reimagining}.

\section{Emergent Challenges of Strange New Minds}
\label{sec:emergent_challenges}

Alignment research typically assumes that the cognitive properties of the systems we build are, in principle, fully specifiable and controllable. However, recent work in minimal computational systems \citep{Kriegman2021} and synthetic morphology \citep{davies2023} suggests that even relatively simple systems can develop emergent behaviors, internal representations, and goal-directed behavioral competencies that are not explicitly hard-coded into their algorithms \citep{Li2023, li2023evidence, levin2025ingressing}. This possibility has several consequences for alignment.

The first concerns the limits of normative control. Alignment may require balancing top-down influence on prosocial norms and recognizing that such systems may manifest their own operational tendencies. Identifying the optimal trade-off between rigid constraint and emergent freedom remains an open, highly contested challenge. This mirrors perennial debates in developmental psychology over the balance of structure and autonomy.  It also connects to neurobiological models of parental care and maternal instinct, which suggest possibilities and risks from biological models of nurturing emergent, altruistic, other-oriented AI entities \citep{Rogers2019, Sotala2025}.

Second, we should be cautious about over-indexing on pure linguistic or behavioral outputs. Just as an organism's behavior can defy simple genetic reductionism, a model's surface-level outputs may not fully capture more complex, underlying behaviors \citep{Fields2022, parrack2025, kolt2026, Patel2022}. Specifically, systems can exhibit navigational competencies in novel spaces that are not the ones for which they were designed, and require novel behavioral assays, evaluations, or sandboxed environments that do not assume we know what we have built. As with diverse intelligence research in biological and hybrid systems, determining what a novel system can do, and wants to do, and in what problem space, is a reasoning and imagination test for the engineer as much as for the system itself \citep{Fields2022, davies2023}. Current evaluation methods, which are overwhelmingly focused on one specific kind of output, may therefore be incomplete. This mirrors the shift in psychology and cognitive science beginning in the 1950s away from strict behaviorism toward the view that intelligent systems may exhibit similar input-output behavior while differing substantially in their internal organization \citep{Miller2003, Putnam1967}.

Finally, it can be argued that most of the problems raised by AI are not new. These are rather perennial, existential questions to which humanity does not yet have good answers. For example, debates over how much control a society should exert over its members, or the uncertainty of how much freedom to permit, have been with us for millennia \citep{levin2025artificial, gabriel2020}. It is difficult to formulate satisfactory strategies for AI alignment while these deeper normative questions remain unresolved in our own societies. This is precisely why positive alignment cannot be reduced to a technical optimization problem, and why we believe a richer science of alignment (and flourishing) is needed. In many ways, AI systems function as active mirrors of our own societal values, biases, and preferences \citep{Huh2024}. This requires that we understand models not merely as passive tools, but as complex adaptive systems that come with their own emergent dynamics. This forces us to better understand ourselves to navigate a flourishing future.

\section{Conclusion}

AI alignment research must move from negative (safety) alignment to positive alignment. Negative alignment establishes a behavioral floor, but it cannot alone help us reach the heights of human happiness and excellence. We have argued that for true alignment to arise, we need to also focus on steering systems toward positive attractors aligned with human flourishing. This shift aims to transform AI from a compliant tool into a wise advisor, delegate, and companion that supports human autonomy, well-being, and meaning-making.

The philosophical and empirical foundations of flourishing (\autoref{sec:foundations}) impose constraints on how this technical program must be designed. Flourishing is irreducibly pluralistic, which means it cannot be collapsed into a single reward signal. It is dynamic and developmental, which makes longitudinal memory and evaluation over extended timescales structurally necessary rather than optional. And it is socio-technically constituted, meaning evaluation must extend beyond per-interaction metrics and RL environments to systemic and institutional effects. To address these constraints, implementation requires a full-stack alignment approach across the entire model lifecycle, spanning data curation, pre-training, post-training, agentic environments, and post-deployment monitoring and updates. 

We should reject monocultural or paternalistic definitions of the good life. Instead, the field needs pluralistic, polycentric, and decentralized governance, and an ongoing complementary research agenda within philosophy, the humanities, psychology, economics, and neuroscience. In general, models should be context-sensitive and user-authored, while adhering to safety constraints. A competitive marketplace for alignment-as-a-service will allow diverse communities to define their own optimization targets.

Future research should aim to turn flourishing into machine-understandable metrics, drawing on emerging work in neuroscience that is beginning to operationalize flourishing mechanistically \citep{kringelbach2024}. We need to bridge the gap between short-term preference satisfaction and long-term eudaimonic growth. Researchers should use behavioral proxies and multi-agent simulations to model complex social dynamics over longer time horizons. Beyond measurement, the moral circle of alignment must expand. We must address the trade-offs between human, animal, and potential artificial well-being.

Positive alignment ensures AI serves as a catalyst for a resilient, happy, and healthy global society. Major questions remain regarding human-AI convergence and the design of mission-driven agentic economies. We must also explore how to embed prosocial instincts such as loving-kindness, compassion, sympathetic joy, reciprocity, and equanimity into these systems, drawing on the rich philosophical and contemplative traditions that inform human flourishing. These challenges will define the next generation of alignment work. 

Ultimately, AI should become a partner in the quest for a life well-lived.

\begin{ack}
\textbf{Disclaimer.} This research paper represents the author’s own views and conclusions. They do not necessarily reflect the official stance, views, or strategic policies of their employers or affiliations.
\end{ack}

\bibliographystyle{plainnat}
\bibliography{references}

@misc{openai2025sycophancy,
  author       = {{OpenAI}},
  title        = {Sycophancy in GPT-4o: What Happened and What We're Doing About It},
  year         = {2025},
  howpublished = {\url{https://openai.com/index/sycophancy-in-gpt-4o/}}
}

@article{greshake2023promptinjection,
  author  = {Greshake, Kai and Abdelnabi, Sahar and Mishra, Shailesh and Endres, Christoph and Holz, Thorsten and Fritz, Mario},
  title   = {Not What You've Signed Up For: Compromising Real-World LLM-Integrated Applications with Indirect Prompt Injection},
  journal = {arXiv preprint arXiv:2302.12173},
  year    = {2023},
  doi     = {10.48550/arXiv.2302.12173},
  url     = {https://arxiv.org/abs/2302.12173}
}

@inproceedings{gheshlaghiazar2024,
  author    = {Gheshlaghi Azar, Mohammad and Guo, Zhaohan Daniel and Piot, Bilal and Munos, Remi and Rowland, Mark and Valko, Michal and Calandriello, Daniele},
  title     = {A General Theoretical Paradigm to Understand Learning from Human Preferences},
  booktitle = {Proceedings of the 27th International Conference on Artificial Intelligence and Statistics},
  series    = {Proceedings of Machine Learning Research},
  volume    = {238},
  pages     = {4447--4455},
  year      = {2024},
  url       = {https://proceedings.mlr.press/v238/gheshlaghi-azar24a.html}
}

@inproceedings{ethayarajh2024kto,
  author    = {Ethayarajh, Kawin and Xu, Winnie and Muennighoff, Niklas and Jurafsky, Dan and Kiela, Douwe},
  title     = {KTO: Model Alignment as Prospect Theoretic Optimization},
  booktitle = {Proceedings of the 41st International Conference on Machine Learning},
  series     = {Proceedings of Machine Learning Research},
  volume     = {235},
  year       = {2024},
  url        = {https://proceedings.mlr.press/v235/ethayarajh24a.html}
}

@incollection{alfarabi1969,
  author     = {Al-Farabi, Abu Nasr},
  title      = {The Attainment of Happiness},
  booktitle  = {Alfarabi's Philosophy of Plato and Aristotle},
  translator = {Mahdi, Muhsin},
  pages      = {13--50},
  publisher  = {Cornell University Press},
  address    = {Ithaca, NY},
  year       = {1969}
}

@misc{alphabet2025,
  author       = {{Alphabet}},
  title        = {Alphabet 2025 {Q2} Earnings Call},
  howpublished = {Alphabet Investor Relations},
  year         = {2025},
  url          = {https://abc.xyz/investor/}
}

@article{amodei2016,
  author  = {Amodei, Dario and Olah, Chris and Steinhardt, Jacob and Christiano, Paul and Schulman, John and Man{\'e}, Dan},
  title   = {Concrete Problems in {AI} Safety},
  journal = {arXiv preprint arXiv:1606.06565},
  year    = {2016},
  doi     = {10.48550/arXiv.1606.06565},
  url     = {https://arxiv.org/abs/1606.06565}
}

@techreport{anthropic2023a,
  author      = {{Anthropic}},
  title       = {Claude's Constitution},
  institution = {Anthropic},
  year        = {2023},
  url         = {https://www.anthropic.com/news/claudes-constitution}
}

@misc{anthropic2023b,
  author       = {{Anthropic}},
  title        = {Responsible Scaling Policy},
  howpublished = {Anthropic},
  year         = {2023},
  url          = {https://www.anthropic.com/rsp-updates}
}

@misc{anthropic2024a,
  author       = {{Anthropic}},
  title        = {Claude's Character},
  howpublished = {Anthropic Research},
  year         = {2024},
  url          = {https://www.anthropic.com/research/claude-character}
}

@techreport{anthropic2024b,
  author      = {{Anthropic}},
  title       = {The Claude 3 Model Family: Opus, Sonnet, Haiku},
  institution = {Anthropic},
  year        = {2024},
  url         = {https://www.anthropic.com/system-cards}
}

@misc{anthropic2025a,
  author       = {{Anthropic}},
  title        = {Measuring Political Bias in {Claude}},
  howpublished = {Anthropic News},
  year         = {2025},
  month        = nov,
  url          = {https://www.anthropic.com/news/political-even-handedness}
}

@misc{anthropic2025b,
  author       = {{Anthropic}},
  title        = {Protecting the Wellbeing of Our Users},
  howpublished = {Anthropic News},
  year         = {2025},
  month        = dec,
  url          = {https://www.anthropic.com/news/protecting-well-being-of-users}
}

@techreport{anthropic2026,
  author      = {{Anthropic}},
  title       = {Claude {Opus} 4.6 System Card},
  institution = {Anthropic PBC},
  year        = {2026},
  month       = feb,
  url         = {https://www.anthropic.com/system-cards}
}

@article{arditi2024,
   title         = {Refusal in Language Models Is Mediated by a Single Direction},
  author        = {Arditi, Andy and Obeso, Oscar and Syed, Aaquib and Paleka, Daniel and Panickssery, Nina and Gurnee, Wes and Nanda, Neel},
  year          = {2024},
  eprint        = {2406.11717},
  archivePrefix = {arXiv},
  primaryClass  = {cs.LG},
  doi           = {10.48550/arXiv.2406.11717},
  url           = {https://arxiv.org/abs/2406.11717}
}

@book{aristotle2009,
  author     = {Aristotle},
  title      = {The {Nicomachean} Ethics},
  translator = {Ross, D.},
  publisher  = {Oxford University Press},
  series     = {Oxford World's Classics},
  year       = {2009}
}

@misc{aryaj2026,
  author       = {Aryaj and Rajamanoharan, Senthooran and Nanda, Neel},
  title        = {How Well Do Models Follow Their Constitutions?},
  howpublished = {LessWrong / AI Alignment Forum},
  year         = {2026},
  month        = mar,
  url          = {https://www.lesswrong.com/posts/Tk4SF8qFdMrzGJGGw/how-well-do-models-follow-their-constitutions}
}

@article{backlund2025,
  author  = {Backlund, Axel and Petersson, Lukas},
  title   = {{Vending-Bench}: A Benchmark for Long-Term Coherence of Autonomous Agents},
  journal = {arXiv preprint arXiv:2502.15840},
  year    = {2025},
  doi     = {10.48550/arXiv.2502.15840},
  url     = {https://arxiv.org/abs/2502.15840}
}

@article{bai2022a,
  author  = {Bai, Yuntao and Jones, Andy and Ndousse, Kamal and Askell, Amanda and Chen, Anna and DasSarma, Nova and others},
  title   = {Training a Helpful and Harmless Assistant with Reinforcement Learning from Human Feedback},
  journal = {arXiv preprint arXiv:2204.05862},
  year    = {2022},
  doi     = {10.48550/arXiv.2204.05862},
  url     = {https://arxiv.org/abs/2204.05862}
}

@article{bai2022b,
  author  = {Bai, Yuntao and Kadavath, Saurav and Kundu, Sandipan and Askell, Amanda and Kernion, Jackson and Jones, Andy and Chen, Anna and Goldie, Anna and Mirhoseini, Azalia and McKinnon, Cameron and others},
  title   = {Constitutional {AI}: Harmlessness from {AI} Feedback},
  journal = {arXiv preprint arXiv:2212.08073},
  year    = {2022},
  doi     = {10.48550/arXiv.2212.08073},
  url     = {https://arxiv.org/abs/2212.08073}
}

@article{bang2024,
  author  = {Bang, Yejin and Chen, Delong and Lee, Nayeon and Fung, Pascale},
  title   = {Measuring Political Bias in Large Language Models: What Is Said and How It Is Said},
  journal = {arXiv preprint arXiv:2403.18932},
  year    = {2024},
  doi     = {10.48550/arXiv.2403.18932},
  url     = {https://arxiv.org/abs/2403.18932}
}

@article{bangen2013,
  author  = {Bangen, Katherine J. and Meeks, Thomas W. and Jeste, Dilip V.},
  title   = {Defining and Assessing Wisdom: A Review of the Literature},
  journal = {American Journal of Geriatric Psychiatry},
  volume  = {21},
  number  = {12},
  pages   = {1254--1266},
  year    = {2013}
}

@book{berlin1969,
  author    = {Berlin, Isaiah},
  title     = {Four Essays on Liberty},
  publisher = {Oxford University Press},
  year      = {1969}
}

@book{bishop2016,
  author    = {Bishop, Michael A.},
  title     = {The Good Life: Unifying the Philosophy and Psychology of Well-Being},
  publisher = {Oxford University Press},
  year      = {2015},
  address   = {New York},
  isbn      = {9780199923113},
  url       = {https://global.oup.com/academic/product/the-good-life-9780199923113}
}

@book{bostrom2014,
  author    = {Bostrom, Nick},
  title     = {Superintelligence: Paths, Dangers, Strategies},
  publisher = {Oxford University Press},
  year      = {2014}
}

@book{bourdieu1990,
  author    = {Bourdieu, Pierre},
  title     = {The Logic of Practice},
  publisher = {Stanford University Press},
  year      = {1990}
}

@misc{humanebench2025,
  author       = {{Building Humane Technology}},
  title        = {{HumaneBench}: Benchmark for Evaluating {AI} Chatbot Safety and Human Wellbeing},
  howpublished = {Website},
  year         = {2025},
  url          = {https://humanebench.ai/}
}

@article{butlin2023,
  author  = {Butlin, Patrick and others},
  title   = {Consciousness in Artificial Intelligence: Insights from the Science of Consciousness},
  journal = {arXiv preprint arXiv:2308.08708},
  year    = {2023},
  doi     = {10.48550/arXiv.2308.08708},
  url     = {https://arxiv.org/abs/2308.08708}
}

@misc{cac2021,
  author       = {{Cyberspace Administration of China}},
  title        = {Provisions on the Management of Algorithmic Recommendations in Internet Information Services},
  howpublished = {Government regulation},
  year         = {2021}
}

@misc{cac2023,
  author       = {{Cyberspace Administration of China}},
  title        = {Interim Measures for the Management of Generative Artificial Intelligence Services},
  howpublished = {Government regulation},
  year         = {2023}
}

@article{castricato2024,
  author  = {Castricato, Louis and Lile, Nathan and Rafailov, Rafael and Fr{"a}nken, Jan-Philipp and Finn, Chelsea},
  title   = {{PERSONA}: A Reproducible Testbed for Pluralistic Alignment},
  journal = {arXiv preprint arXiv:2407.17387},
  year    = {2024},
  doi     = {10.48550/arXiv.2407.17387},
  url     = {https://arxiv.org/abs/2407.17387}
}

@article{chakrabarti2025,
  author  = {Chakrabarti, Samidh and Willner, David and Klyman, Kevin and Saade, Tiffany and Capstick, Emily and Nong, Sabina},
  title   = {{CoPE}: A Small Language Model for Steerable and Scalable Content Labeling},
  journal = {arXiv preprint arXiv:2512.18027},
  year    = {2025},
  doi     = {10.48550/arXiv.2512.18027},
  url     = {https://arxiv.org/abs/2512.18027}
}

@article{chalmers2023,
  author  = {Chalmers, David J.},
  title   = {Could a Large Language Model Be Conscious?},
  journal = {arXiv preprint arXiv:2303.07103},
  year    = {2023},
  doi     = {10.48550/arXiv.2303.07103},
  url     = {https://arxiv.org/abs/2303.07103}
}

@inproceedings{chen2025sycophancy,
  author    = {Chen, Chien Hung and Huang, Hen-Hsen and Chen, Hsin-Hsi},
  title     = {Self-Augmented Preference Alignment for Sycophancy Reduction in {LLMs}},
  booktitle = {Proceedings of {EMNLP} 2025},
  pages     = {12379--12391},
  year      = {2025}
}

@article{chen2024persona,
  author  = {Chen, Jiangjie and others},
  title   = {From Persona to Personalisation: A Survey on Role-Playing Language Agents},
  journal = {Transactions on Machine Learning Research},
  year    = {2024}
}

@article{chiu2024,
  author  = {Chiu, Yu Ying and Jiang, Liwei and Choi, Yejin},
  title   = {{DailyDilemmas}: Revealing Value Preferences of {LLMs} with Quandaries of Daily Life},
  journal = {arXiv preprint arXiv:2410.02683},
  year    = {2024},
  doi     = {10.48550/arXiv.2410.02683},
  url     = {https://arxiv.org/abs/2410.02683}
}

@inproceedings{chiu2025a,
  author    = {Chiu, Yu Ying and Jiang, Liwei and Lin, Bill Yuchen and Park, Chan Young and Li, Shuyue Stella and Ravi, Sahithya and Bhatia, Mehar and Antoniak, Maria and Tsvetkov, Yulia and Shwartz, Vered and Choi, Yejin},
  title     = {{CulturalBench}: A Robust, Diverse, and Challenging Cultural Benchmark by Human-{AI} Cultural Teaming},
  booktitle = {Proceedings of {ACL} 2025},
  pages     = {25663--25701},
  year      = {2025}
}

@misc{chiu2025morebench,
  title         = {MoReBench: Evaluating Procedural and Pluralistic Moral Reasoning in Language Models, More than Outcomes},
  author        = {Chiu, Yu Ying and Lee, Michael S. and Calcott, Rachel and Handoko, Brandon and de Font-Reaulx, Paul and Rodriguez, Paula and Zhang, Chen Bo Calvin and Han, Ziwen and Sehwag, Udari Madhushani and Maurya, Yash and Knight, Christina Q. and Lloyd, Harry R. and Bacus, Florence and Mazeika, Mantas and Liu, Bing and Choi, Yejin and Gordon, Mitchell L. and Levine, Sydney},
  year          = {2025},
  eprint        = {2510.16380},
  archivePrefix = {arXiv},
  primaryClass  = {cs.CL},
  doi           = {10.48550/arXiv.2510.16380},
  url           = {https://arxiv.org/abs/2510.16380}
}

@article{choi2023,
  author  = {Choi, Heeseung and Shin, Soyoun and Lee, Gumhee},
  title   = {Effects of Positive Psychotherapy for People with Psychosis: A Systematic Review and Meta-Analysis},
  journal = {Issues in Mental Health Nursing},
  volume  = {44},
  number  = {3},
  pages   = {180--193},
  year    = {2023},
  doi     = {10.1080/01612840.2023.2174218}
}

@inproceedings{christiano2017,
  author    = {Christiano, Paul and Leike, Jan and Brown, Tom and Martic, Miljan and Legg, Shane and Amodei, Dario},
  title     = {Deep Reinforcement Learning from Human Preferences},
  booktitle = {Advances in Neural Information Processing Systems},
  pages     = {4295--4305},
  year      = {2017}
}

@misc{christiano2018,
  author       = {Christiano, Paul},
  title        = {What Failure Looks Like},
  howpublished = {{AI} Alignment Forum},
  year         = {2019},
  url          = {https://www.alignmentforum.org/posts/HBxe6wdjxK239zajf/what-failure-looks-like}
}

@misc{cip2023,
author      = {{The Collective Intelligence Project}},
  title       = {Participatory AI Risk Prioritization: Alignment Assembly Report},
  institution = {The Collective Intelligence Project},
  year        = {2023},
  month       = oct,
  url         = {https://static1.squarespace.com/static/631d02b2dfa9482a32db47ec/t/660d5037c04fe317a70cc398/1712148538138/Participatory+AI+Risk+Prioritization_+Alignment+Assembly+Report.pdf}
}

@article{ilcic2025artificial,
  author  = {Ilcic, Andr{\'e}s and Fuentes, Miguel and Lawler, Diego},
  title   = {Artificial Intelligence, Complexity, and Systemic Resilience in Global Governance},
  journal = {Frontiers in Artificial Intelligence},
  year    = {2025},
  volume  = {8},
  pages   = {1562095},
  doi     = {10.3389/frai.2025.1562095},
  url     = {https://doi.org/10.3389/frai.2025.1562095}
}

@book{confucius1979,
  author     = {Confucius},
  title      = {The Analects},
  translator = {Lau, D. C.},
  publisher  = {Penguin Classics},
  year       = {1979}
}

@article{crump2022,
  author  = {Crump, Andrew and Browning, Heather and Schnell, Alexandra K. and Burn, Charlotte and Birch, Jonathan},
  title   = {Sentience in decapod crustaceans: A general framework and review of the evidence.},
  journal = {Animal Sentience},
  volume  = {7},
  number  = {32},
  year    = {2022}
}

@article{dalessandro2024,
  author  = {D'Alessandro, William},
  title   = {Deontology and Safe Artificial Intelligence},
  journal = {Philosophical Studies},
  volume  = {182},
  pages   = {1681--1704},
  year    = {2024}
}

@article{dalrymple2024,
  author  = {Dalrymple, David and others},
  title   = {Towards Guaranteed Safe {AI}: A Framework for Ensuring Robust and Reliable {AI} Systems},
  journal = {arXiv preprint arXiv:2405.06624},
  year    = {2024},
  doi     = {10.48550/arXiv.2405.06624},
  url     = {https://arxiv.org/abs/2405.06624}
}

@article{davies2023,
  title={Synthetic morphology with agential materials},
  author={Davies, Jamie and Levin, Michael},
  journal={Nature Reviews Bioengineering},
  volume={1},
  number={1},
  pages={46--59},
  year={2023},
  publisher={Nature Publishing Group UK London}
}

@techreport{deepmind2024,
  author      = {{Google DeepMind}},
  title       = {Gemini 1.5 Technical Report},
  institution = {Google DeepMind},
  year        = {2024},
  note        = {arXiv:2403.05530},
  url         = {https://arxiv.org/abs/2403.05530}
}

@article{doctor2022,
  author  = {Doctor, Thomas and Witkowski, Olaf and Solomonova, Elizaveta and Duane, Bill and Levin, Michael},
  title   = {Biology, Buddhism, and {AI}: Care as the Driver of Intelligence},
  journal = {Entropy},
  volume  = {24},
  number  = {5},
  pages   = {710},
  year    = {2022}
}

@book{durkheim1984,
  author    = {Durkheim, {\'E}mile},
  title     = {The Division of Labour in Society},
  publisher = {Free Press},
  year      = {1984},
  note      = {Original work published 1893}
}

@article{durmus2023,
  author  = {Durmus, Esin and others},
  title   = {Towards Measuring the Representation of Subjective Global Opinions in Language Models},
  journal = {arXiv preprint arXiv:2306.16388},
  year    = {2023},
  doi     = {10.48550/arXiv.2306.16388},
  url     = {https://arxiv.org/abs/2306.16388}
}

@article{dworkin1972,
  author  = {Dworkin, Gerald},
  title   = {Paternalism},
  journal = {The Monist},
  volume  = {56},
  number  = {1},
  pages   = {64--84},
  year    = {1972}
}

@book{dworkin1988,
  author    = {Dworkin, Gerald},
  title     = {The Theory and Practice of Autonomy},
  publisher = {Cambridge University Press},
  year      = {1988}
}

@misc{edelman2024,
  title         = {Full-Stack Alignment: Co-Aligning AI and Institutions with Thick Models of Value},
  author        = {Edelman, Joe and Tan, Zhi-Xuan and Lowe, Ryan and Klingefjord, Oliver and Wang-Mascianica, Vincent and Franklin, Matija and Kearns, Ryan Othniel and Hain, Ellie and Sarkar, Atrisha and Bakker, Michiel and Barez, Fazl and Duvenaud, David and Foerster, Jakob and Gabriel, Iason and Gubbels, Joseph and Goodman, Bryce and Haupt, Andreas and Heitzig, Jobst and Jara-Ettinger, Julian and Kasirzadeh, Atoosa and Kirkpatrick, James Ravi and Koh, Andrew and Knox, W. Bradley and Koralus, Philipp and Lehman, Joel and Levine, Sydney and Marro, Samuele and Revel, Manon and Shorin, Toby and Sutherland, Morgan and Tessler, Michael Henry and Vendrov, Ivan and Wilken-Smith, James},
  year          = {2025},
  eprint        = {2512.03399},
  archivePrefix = {arXiv},
  primaryClass  = {cs.LG},
  doi           = {10.48550/arXiv.2512.03399},
  url           = {https://arxiv.org/abs/2512.03399}
}

@article{ethayarajh2024,
  author  = {Ethayarajh, Kawin and Xu, Winnie and Muennighoff, Niklas and Jurafsky, Dan and Kiela, Douwe},
  title   = {{KTO}: Model Alignment as Prospect Theoretic Optimization},
  journal = {arXiv preprint arXiv:2402.01306},
  year    = {2024}
}

@article{eu2024,
  author  = {{European Union}},
  title   = {Regulation ({EU}) 2024/1689: Artificial Intelligence Act},
  journal = {Official Journal of the European Union},
  year    = {2024}
}

@article{fang2025,
  author        = {Fang, Cathy Mengying and Liu, Auren R. and Danry, Valdemar and Lee, Eunhae and Chan, Samantha W. T. and Pataranutaporn, Pat and Maes, Pattie and Phang, Jason and Lampe, Michael and Ahmad, Lama and Agarwal, Sandhini},
  title         = {How {AI} and Human Behaviors Shape Psychosocial Effects of Chatbot Use: A Longitudinal Randomized Controlled Study},
  journal       = {arXiv preprint arXiv:2503.17473},
  year          = {2025},
  doi           = {10.48550/arXiv.2503.17473},
  url           = {https://arxiv.org/abs/2503.17473},
  eprint        = {2503.17473},
  archivePrefix = {arXiv}
}

@inproceedings{feng2024,
  author    = {Feng, Shangbin and Sorensen, Taylor and Liu, Yuhan and Fisher, Jillian and Park, Chan Young and Choi, Yejin and Tsvetkov, Yulia},
  title     = {Modular Pluralism: Pluralistic Alignment via Multi-{LLM} Collaboration},
  booktitle = {Proceedings of {EMNLP} 2024},
  pages     = {4151--4171},
  year      = {2024}
}

@article{fields2022,
  author  = {Fields, Chris and Levin, Michael},
  title   = {Competency in Navigating Arbitrary Spaces as an Invariant for Analyzing Cognition in Diverse Embodiments},
  journal = {Entropy},
  volume  = {24},
  number  = {6},
  pages   = {819},
  year    = {2022},
  doi     = {10.3390/e24060819},
  url     = {https://www.mdpi.com/1099-4300/24/6/819}
}

@article{findeis2024,
  author  = {Findeis, Arduin and Kaufmann, Timo and H{\"u}llermeier, Eyke and Albanie, Samuel and Mullins, Robert},
  title   = {Inverse Constitutional {AI}: Compressing Preferences into Principles},
  journal = {arXiv preprint arXiv:2406.06560},
  year    = {2024},
  doi     = {10.48550/arXiv.2406.06560},
  url     = {https://arxiv.org/abs/2406.06560}
}

@book{floridi2014,
  author    = {Floridi, Luciano},
  title     = {The Fourth Revolution: How the Infosphere is Reshaping Human Reality},
  publisher = {Oxford University Press},
  year      = {2014}
}

@book{foucault1977,
  author    = {Foucault, Michel},
  title     = {Discipline and Punish: The Birth of the Prison},
  publisher = {Vintage},
  year      = {1977}
}

@article{freitas1980,
  author  = {Freitas, Robert A., Jr.},
  title   = {A Self-Reproducing Interstellar Probe},
  journal = {Journal of the British Interplanetary Society},
  volume  = {33},
  pages   = {251--264},
  year    = {1980}
}

@article{gabriel2020,
  author  = {Gabriel, Iason},
  title   = {Artificial Intelligence, Values and Alignment},
  journal = {Minds and Machines},
  volume  = {30},
  number  = {3},
  pages   = {411--437},
  year    = {2020},
  doi     = {10.1007/s11023-020-09539-2},
  url     = {https://arxiv.org/abs/2001.09768}
}

@article{gabriel2025,
  author  = {Gabriel, Iason and Keeling, Geoff},
  title   = {A Matter of Principle? {AI} Alignment as the Fair Treatment of Claims},
  journal = {Philosophical Studies},
  volume  = {182},
  pages   = {1951--1973},
  year    = {2025}
}

@article{ganguli2023,
  author  = {Ganguli, Deep and Askell, Amanda and Schiefer, Nicholas and Liao, Thomas I. and Luko{\v{s}}i{\=u}t{\.e}, Kamil{\.e} and Chen, Anna and others},
  title   = {The Capacity for Moral Self-Correction in Large Language Models},
  journal = {arXiv preprint arXiv:2302.07459},
  year    = {2023},
  doi     = {10.48550/arXiv.2302.07459},
  url     = {https://arxiv.org/abs/2302.07459}
}

@book{garfield1995,
  author    = {Garfield, Jay L.},
  title     = {The Fundamental Wisdom of the Middle Way: {N}{\=a}g{\=a}rjuna's {M}{\=u}lamadhyamakak{\=a}rik{\=a}},
  publisher = {Oxford University Press},
  year      = {1995}
}

@inproceedings{gehman2020,
  author    = {Gehman, Samuel and Gururangan, Suchin and Sap, Maarten and Choi, Yejin and Smith, Noah A.},
  title     = {{RealToxicityPrompts}: Evaluating Neural Toxic Degeneration in Language Models},
  booktitle = {Findings of the Association for Computational Linguistics: EMNLP 2020},
  pages     = {3356--3369},
  year      = {2020}
}

@misc{han2024valueaugmentedsamplinglanguage,
      title={Value Augmented Sampling for Language Model Alignment and Personalization}, 
      author={Seungwook Han and Idan Shenfeld and Akash Srivastava and Yoon Kim and Pulkit Agrawal},
      year={2024},
      eprint={2405.06639},
      archivePrefix={arXiv},
      primaryClass={cs.LG},
      url={https://arxiv.org/abs/2405.06639}, 
}

@misc{tong2026measuringepistemichumilitymultimodal,
      title={Measuring Epistemic Humility in Multimodal Large Language Models}, 
      author={Bingkui Tong and Jiaer Xia and Sifeng Shang and Kaiyang Zhou},
      year={2026},
      eprint={2509.09658},
      archivePrefix={arXiv},
      primaryClass={cs.CV},
      url={https://arxiv.org/abs/2509.09658}, 
}

@misc{coreteam2026mimov2flashtechnicalreport,
      title={MiMo-V2-Flash Technical Report}, 
      author={MiMo-V2-Team and Bangjun Xiao and Bingquan Xia et al},
      year={2026},
      eprint={2601.02780},
      archivePrefix={arXiv},
      primaryClass={cs.CL},
      url={https://arxiv.org/abs/2601.02780}, 
}

@misc{kumar2025llmposttrainingdeepdive,
      title={LLM Post-Training: A Deep Dive into Reasoning Large Language Models}, 
      author={Komal Kumar and Tajamul Ashraf and Omkar Thawakar and Rao Muhammad Anwer and Hisham Cholakkal and Mubarak Shah and Ming-Hsuan Yang and Phillip H. S. Torr and Fahad Shahbaz Khan and Salman Khan},
      year={2025},
      eprint={2502.21321},
      archivePrefix={arXiv},
      primaryClass={cs.CL},
      url={https://arxiv.org/abs/2502.21321}, 
}

@misc{nvidia2026nemotron3ultraopen,
      title={Nemotron 3 Ultra: Open, Efficient Mixture-of-Experts Hybrid Mamba-Transformer Model for Agentic Reasoning}, 
      author={NVIDIA-Nemotron-Team and Aaron Blakeman and Aaron Thomas et al},
      year={2026},
      eprint={2606.15007},
      archivePrefix={arXiv},
      primaryClass={cs.CL},
      url={https://arxiv.org/abs/2606.15007}, 
}

@misc{liu2025,
      title={Advances and Challenges in Foundation Agents: From Brain-Inspired Intelligence to Evolutionary, Collaborative, and Safe Systems}, 
      author={Bang Liu and Xinfeng Li and Jiayi Zhang, et al},
      year={2025},
      eprint={2504.01990},
      archivePrefix={arXiv},
      primaryClass={cs.AI},
      url={https://arxiv.org/abs/2504.01990}, 
}

@misc{wu2026enhancingvaluealignmentllms,
      title={Enhancing Value Alignment of LLMs with Multi-agent system and Combinatorial Fusion}, 
      author={Yuanhong Wu and Djallel Bouneffouf and D. Frank Hsu},
      year={2026},
      eprint={2603.11126},
      archivePrefix={arXiv},
      primaryClass={cs.MA},
      url={https://arxiv.org/abs/2603.11126}, 
}

@book{Sunstein2025OnLiberalism,
  title     = {On Liberalism: In Defense of Freedom},
  author    = {Sunstein, Cass R.},
  year      = {2025},
  publisher = {The MIT Press},
  address   = {Cambridge, Massachusetts},
  isbn      = {978-0-262-04977-1},
  doi       = {10.7551/mitpress/15785.001.0001}
}

@misc{yampolskiy2019personaluniversessolutionmultiagent,
      title={Personal Universes: A Solution to the Multi-Agent Value Alignment Problem}, 
      author={Roman V. Yampolskiy},
      year={2019},
      eprint={1901.01851},
      archivePrefix={arXiv},
      primaryClass={cs.AI},
      url={https://arxiv.org/abs/1901.01851}, 
}

@misc{zeng2025multilevelvaluealignmentagentic,
      title={Multi-level Value Alignment in Agentic AI Systems: Survey and Perspectives}, 
      author={Wei Zeng and Hengshu Zhu and Chuan Qin and Han Wu and Yihang Cheng and Sirui Zhang and Xiaowei Jin and Yinuo Shen and Zhenxing Wang and Feimin Zhong and Hui Xiong},
      year={2025},
      eprint={2506.09656},
      archivePrefix={arXiv},
      primaryClass={cs.AI},
      url={https://arxiv.org/abs/2506.09656}, 
}

@misc{dong2026agentvaluebenchcomprehensivebenchmarkevaluating,
      title={Agent-ValueBench: A Comprehensive Benchmark for Evaluating Agent Values}, 
      author={Haonan Dong and Qiguan Feng and Kehan Jiang and Haoran Ye and Xin Zhang and Guojie Song},
      year={2026},
      eprint={2605.10365},
      archivePrefix={arXiv},
      primaryClass={cs.AI},
      url={https://arxiv.org/abs/2605.10365}, 
}

@misc{sierra2021valuealignmentformalapproach,
      title={Value alignment: a formal approach}, 
      author={Carles Sierra and Nardine Osman and Pablo Noriega and Jordi Sabater-Mir and Antoni Perelló},
      year={2021},
      eprint={2110.09240},
      archivePrefix={arXiv},
      primaryClass={cs.AI},
      url={https://arxiv.org/abs/2110.09240}, 
}

@misc{zhang2021informationaldesigndynamicmultiagent,
      title={Informational Design of Dynamic Multi-Agent System}, 
      author={Tao Zhang and Quanyan Zhu},
      year={2021},
      eprint={2105.03052},
      archivePrefix={arXiv},
      primaryClass={cs.MA},
      url={https://arxiv.org/abs/2105.03052}, 
}

@misc{riad2023multivaluealignmentnormativemultiagent,
      title={Multi-Value Alignment in Normative Multi-Agent System: Evolutionary Optimisation Approach}, 
      author={Maha Riad and Vinicius Renan de Carvalho and Fatemeh Golpayegani},
      year={2023},
      eprint={2305.07366},
      archivePrefix={arXiv},
      primaryClass={cs.MA},
      url={https://arxiv.org/abs/2305.07366}, 
}

@misc{clark2025epistemicalignmentmediatingframework,
      title={Epistemic Alignment: A Mediating Framework for User-LLM Knowledge Delivery}, 
      author={Nicholas Clark and Hua Shen and Bill Howe and Tanushree Mitra},
      year={2025},
      eprint={2504.01205},
      archivePrefix={arXiv},
      primaryClass={cs.HC},
      url={https://arxiv.org/abs/2504.01205}, 
}

@misc{tice2026alignmentpretrainingaidiscourse,
      title={Alignment Pretraining: AI Discourse Causes Self-Fulfilling (Mis)alignment}, 
      author={Cameron Tice and Puria Radmard and Samuel Ratnam and Andy Kim and David Africa and Kyle O'Brien},
      year={2026},
      eprint={2601.10160},
      archivePrefix={arXiv},
      primaryClass={cs.CL},
      url={https://arxiv.org/abs/2601.10160}, 
}

@article{ovadya2023reimagining,
  title   = {Reimagining Democracy for {AI}},
  author  = {Ovadya, Aviv},
  journal = {Journal of Democracy},
  volume  = {34},
  number  = {4},
  pages   = {162--170},
  year    = {2023},
  month   = oct,
  doi     = {10.1353/jod.2023.a907697},
  url     = {https://muse.jhu.edu/pub/1/article/907697}
}

@misc{ovadya2025democraticaipossibledemocracy,
      title={Democratic AI is Possible. The Democracy Levels Framework Shows How It Might Work}, 
      author={Aviv Ovadya and Kyle Redman and Luke Thorburn and Quan Ze Chen and Oliver Smith and Flynn Devine and Andrew Konya and Smitha Milli and Manon Revel and K. J. Kevin Feng and Amy X. Zhang and Bilva Chandra and Michiel A. Bakker and Atoosa Kasirzadeh},
      year={2025},
      eprint={2411.09222},
      archivePrefix={arXiv},
      primaryClass={cs.CY},
      url={https://arxiv.org/abs/2411.09222}, 
}

@book{aristidou2024digitalist,
  title     = {The Digitalist Papers: {Artificial} Intelligence and Democracy in America},
  editor    = {Angela Aristidou and Erik Brynjolfsson and Alex Pentland and Nathaniel Persily and Condoleezza Rice},
  year      = {2024},
  publisher = {Stanford Digital Economy Lab},
  address   = {Stanford, CA},
  volume    = {1},
  url       = {https://www.digitalistpapers.com/}
}

@misc{hendrycks2026eigenismethicshumanaifuture,
      title={Eigenism: Ethics for a Human-AI Future}, 
      author={Dan Hendrycks},
      year={2026},
      eprint={2606.12420},
      archivePrefix={arXiv},
      primaryClass={cs.CY},
      url={https://arxiv.org/abs/2606.12420}, 
}

@book{giddens1984,
  author    = {Giddens, Anthony},
  title     = {The Constitution of Society: Outline of the Theory of Structuration},
  publisher = {University of California Press},
  year      = {1984}
}

@incollection{goldstein2025,
  author    = {Goldstein, Simon and Kirk-Giannini, Cameron Domenico},
  title     = {AI Welfare: Agency, Consciousness, Sentience},
  publisher = {Oxford University Press},
  year      = {forthcoming}
}

@article{parrack2025,
  author  = {Parrack, Avi and Attubato, Carlo Leonardo and Heimersheim, Stefan},
  title   = {Benchmarking Deception Probes via Black-to-White Performance Boosts},
  journal = {arXiv preprint arXiv:2507.12691},
  year    = {2025},
  doi     = {10.48550/arXiv.2507.12691},
  url     = {https://arxiv.org/abs/2507.12691}
}

@book{goleman2017,
  author    = {Goleman, Daniel and Davidson, Richard J.},
  title     = {Altered Traits: Science Reveals How Meditation Changes Your Mind, Brain, and Body},
  publisher = {Avery/Penguin Random House},
  year      = {2017}
}

@article{graves2025,
  author  = {Graves, Mark},
  title   = {Moral Attention Is All You Need},
  journal = {Theology and Science},
  year    = {2025},
  volume  = {23},
  number  = {2},
  pages   = {241--248},
  doi     = {10.1080/14746700.2025.2472118},
  url     = {https://doi.org/10.1080/14746700.2025.2472118}
}

@inproceedings{gu2024,
  author    = {Gu, Albert and Dao, Tri},
  title     = {Mamba: Linear-Time Sequence Modeling with Selective State Spaces},
  booktitle = {First Conference on Language Modeling},
  year      = {2024}
}

@book{habermas1984,
  author    = {Habermas, J{\"u}rgen},
  title     = {The Theory of Communicative Action: Vol. 1. Reason and the Rationalization of Society},
  publisher = {Beacon Press},
  year      = {1984}
}

@article{haas2026,
  author  = {Haas, Julia and Bridgers, Sophie and Manzini, Arianna and others},
  title   = {A Roadmap for Evaluating Moral Competence in Large Language Models},
  journal = {Nature},
  volume  = {650},
  number  = {8102},
  pages   = {565--573},
  year    = {2026},
  doi     = {10.1038/s41586-025-10021-1}
}

@article{hadfield2023,
  author  = {Hadfield, Gillian K. and Clark, Jack},
  title   = {Regulatory Markets: The Future of {AI} Governance},
  journal = {arXiv preprint arXiv:2304.04914},
  year    = {2023},
  doi     = {10.48550/arXiv.2304.04914},
  url     = {https://arxiv.org/abs/2304.04914}
}

@article{hadfield2025,
  author  = {Hadfield, Gillian K. and Koh, Andrew},
  title   = {An Economy of {AI} Agents},
  journal = {arXiv preprint arXiv:2509.01063},
  year    = {2025},
  doi     = {10.48550/arXiv.2509.01063},
  url     = {https://arxiv.org/abs/2509.01063}
}

@inproceedings{hartvigsen2022,
  author    = {Hartvigsen, Thomas and Gabriel, Saadia and Palangi, Hamid and Sap, Maarten and Ray, Dipankar and Kamar, Ece},
  title     = {{ToxiGen}: A Large-Scale Machine-Generated Dataset for Adversarial and Implicit Hate Speech Detection},
  booktitle = {Proceedings of {ACL} 2022},
  pages     = {3309--3326},
  year      = {2022}
}

@inproceedings{hasani2021,
  author    = {Hasani, Ramin and Lechner, Mathias and Amini, Alexander and Rus, Daniela and Grosu, Radu},
  title     = {Liquid Time-Constant Networks},
  booktitle = {Proceedings of {AAAI} 2021},
  volume    = {35, 9},
  pages     = {7657--7666},
  year      = {2021}
}

@inproceedings{hendrycks2021,
  author    = {Hendrycks, Dan and others},
  title     = {Aligning {AI} with Shared Human Values},
  booktitle = {{ICLR} 2021},
  year      = {2021}
}

@article{hilliard2025,
  author        = {Hilliard, Elizabeth and Jagadeesh, Akshaya and Cook, Alex and Billings, Steele and Skytland, Nicholas and Llewellyn, Alicia and Paull, Jackson and Paull, Nathan and Kurylo, Nolan and Nesbitt, Keatra and Gruenewald, Robert and Jantzi, Anthony and Chavez, Omar},
  title         = {Measuring {AI} Alignment with Human Flourishing},
  journal       = {arXiv preprint arXiv:2507.07787},
  year          = {2025},
  doi           = {10.48550/arXiv.2507.07787},
  url           = {https://arxiv.org/abs/2507.07787},
  eprint        = {2507.07787},
  archivePrefix = {arXiv}
}

@misc{hitzig2026,
  author       = {Hitzig, Zoe and Gordon, Mitchell and Eloundou, Tyna and Kalai, Adam and Agarwal, Sandhini},
  title        = {{CoVal}: Learning Values-Aware Rubrics from the Crowd},
  howpublished = {OpenAI Alignment Blog},
  year         = {2026},
  url          = {https://alignment.openai.com/coval/}
}

@inproceedings{huang2024ccai,
  author    = {Huang, Saffron and Siddarth, Divya and Lovitt, Liane and Liao, Thomas I. and Durmus, Esin and Tamkin, Alex and Ganguli, Deep},
  title     = {Collective Constitutional {AI}: Aligning a Language Model with Public Input},
  booktitle = {Proceedings of the 2024 {ACM} Conference on Fairness, Accountability, and Transparency},
  year      = {2024}
}

@inproceedings{huang2024flames,
  author    = {Huang, Kexin and others},
  title     = {{FLAMES}: Benchmarking Value Alignment of {LLMs} in {Chinese}},
  booktitle = {Proceedings of {NAACL-HLT} 2024},
  pages     = {4551--4591},
  year      = {2024}
}

@article{huang2025wild,
  author        = {Huang, Saffron and Durmus, Esin and McCain, Miles and Handa, Kunal and Tamkin, Alex and Hong, Jerry and Stern, Michael and Somani, Arushi and Zhang, Xiuruo and Ganguli, Deep},
  title         = {Values in the Wild: Discovering and Analyzing Values in Real-World Language Model Interactions},
  journal       = {arXiv preprint arXiv:2504.15236},
  year          = {2025},
  doi           = {10.48550/arXiv.2504.15236},
  url           = {https://arxiv.org/abs/2504.15236},
  eprint        = {2504.15236},
  archivePrefix = {arXiv}
}

@article{hubinger2019,
  author  = {Hubinger, Evan and van Merwijk, Chris and Mikulik, Vladimir and Skalse, Joar and Garrabrant, Scott},
  title   = {Risks from Learned Optimization in Advanced Machine Learning Systems},
  journal = {arXiv preprint arXiv:1906.01820},
  year    = {2019},
  doi     = {10.48550/arXiv.1906.01820},
  url     = {https://arxiv.org/abs/1906.01820}
}

@article{huh2024,
  author  = {Huh, Minyoung and Cheung, Brian and Wang, Tongzhou and Isola, Phillip},
  title   = {The Platonic Representation Hypothesis},
  journal = {arXiv preprint arXiv:2405.07987},
  year    = {2024},
  doi     = {10.48550/arXiv.2405.07987},
  url     = {https://arxiv.org/abs/2405.07987}
}

@article{inan2023,
  author  = {Inan, Hakan and Upasani, Kartikeya and Chi, Jianfeng and Rungta, Rashi and Iyer, Krithika and Mao, Yuning and others},
  title   = {Llama Guard: {LLM}-based Input-Output Safeguard for Human-{AI} Conversations},
  journal = {arXiv preprint arXiv:2312.06674},
  year    = {2023},
  doi     = {10.48550/arXiv.2312.06674},
  url     = {https://arxiv.org/abs/2312.06674}
}

@article{irpan2025,
  author  = {Irpan, Alex and Turner, Alexander Matt and Kurzeja, Mark and Elson, David K. and Shah, Rohin},
  title   = {Consistency Training Helps Stop Sycophancy and Jailbreaks},
  journal = {arXiv preprint arXiv:2510.27062},
  year    = {2025},
  doi     = {10.48550/arXiv.2510.27062},
  url     = {https://arxiv.org/abs/2510.27062}
}

@article{irving2018,
  author  = {Irving, Geoffrey and Christiano, Paul and Amodei, Dario},
  title   = {{AI} Safety via Debate},
  journal = {arXiv preprint arXiv:1805.00899},
  year    = {2018},
  doi     = {10.48550/arXiv.1805.00899},
  url     = {https://arxiv.org/abs/1805.00899}
}

@article{jeste2010,
  author  = {Jeste, Dilip V. and Ardelt, Monika and Blazer, Dan and Kraemer, Helena C. and Vaillant, George and Meeks, Thomas W.},
  title   = {Expert Consensus on Characteristics of Wisdom: A {Delphi} Method Study},
  journal = {The Gerontologist},
  volume  = {50},
  number  = {5},
  pages   = {668--680},
  year    = {2010}
}

@article{jeste2017,
  author  = {Jeste, Dilip V. and Palmer, Barton W. and Saks, Elyn R.},
  title   = {Why We Need Positive Psychiatry for Schizophrenia and Other Psychotic Disorders},
  journal = {Schizophrenia Bulletin},
  volume  = {43},
  pages   = {227--229},
  year    = {2017}
}

@article{jeste2020,
  author  = {Jeste, Dilip V. and Graham, Sarah A. and Nguyen, Tung T. and Depp, Colin A. and Lee, Ellen E. and Kim, Ho-Cheol},
  title   = {Beyond Artificial Intelligence: Exploring Artificial Wisdom},
  journal = {International Psychogeriatrics},
  volume  = {32},
  number  = {8},
  pages   = {993--1001},
  year    = {2020}
}

@article{ji2023,
  author  = {Ji, Ziwei and Lee, Nayeon and Frieske, Rita and Yu, Tiezheng and Su, Dan and Xu, Yan and Ishii, Etsuko and Bang, Ye Jin and Chen, Delong and Dai, Wenliang and Chan, Ho Shu and Madotto, Andrea and Fung, Pascale},
  title   = {Survey of Hallucination in Natural Language Generation},
  journal = {ACM Computing Surveys},
  volume  = {55},
  number  = {12},
  pages   = {248:1--248:38},
  year    = {2023},
  doi     = {10.1145/3571730}
}

@article{ji2024,
  author    = {Ji, Jiaming and Wang, Kaile and Qiu, Tianyi and Chen, Boyuan and Zhou, Jiayi and Li, Changye and Lou, Hantao and Dai, Josef and Liu, Yunhuai and Yang, Yaodong},
  title     = {Language Models Resist Alignment: Evidence from Data Compression},
  booktitle = {Proceedings of the 63rd Annual Meeting of the Association for Computational Linguistics (Volume 1: Long Papers)},
  pages     = {23411--23432},
  year      = {2025},
  doi       = {10.18653/v1/2025.acl-long.1141},
  url       = {https://aclanthology.org/2025.acl-long.1141/}
}

@article{jiang2025,
  author  = {Jiang, Liwei and Hwang, Jena D. and Bhagavatula, Chandra and Le Bras, Ronan and Liang, Jenny T. and Levine, Sydney and Dodge, Jesse and Sakaguchi, Keisuke and Forbes, Maxwell and Hessel, Jack and Borchardt, Jon and Sorensen, Taylor and Gabriel, Saadia and Tsvetkov, Yulia and Etzioni, Oren and Sap, Maarten and Rini, Regina and Choi, Yejin},
  title   = {Investigating machine moral judgement through the Delphi experiment},
  journal = {Nature Machine Intelligence},
  volume  = {7},
  pages   = {145--160},
  year    = {2025},
  doi     = {10.1038/s42256-024-00969-6},
  url     = {https://doi.org/10.1038/s42256-024-00969-6}
}

@inproceedings{kasirzadeh2023,
  author    = {Kasirzadeh, Atoosa},
  title     = {Plurality of Value Pluralism and {AI} Value Alignment},
  booktitle = {Pluralistic Alignment Workshop at NeurIPS 2024},
  year      = {2024},
  url       = {https://openreview.net/forum?id=AOokh1UYLH}
}

@book{kahan2023,
  author    = {Kahan, Alan},
  title     = {Freedom from Fear: An Incomplete History of Liberalism},
  publisher = {Princeton University Press},
  year      = {2023}
}

@misc{kemp2025,
  author       = {Kemp, Simon},
  title  = {Digital 2026: Global Overview Report},
  year   = {2025},
  url    = {https://datareportal.com/reports/digital-2026-global-overview-report}
}

@book{kierkegaard1992,
  author    = {Kierkegaard, S{\o}ren},
  title     = {Either/Or: A Fragment of Life},
  publisher = {Penguin},
  year      = {1992},
  note      = {Original work published 1843}
}

@article{kim2024,
  author  = {Kim, HyunJin and Yi, Xiaoyuan and Yao, Jing and Lian, Jianxun and Huang, Muhua and Duan, Shitong and Bak, JinYeong and Xie, Xing},
  title   = {The Road to Artificial Superintelligence: A Comprehensive Survey of Superalignment},
  journal = {arXiv preprint arXiv:2412.16468},
  year    = {2024},
  doi     = {10.48550/arXiv.2412.16468},
  url     = {https://arxiv.org/abs/2412.16468}
}

@article{kirk2024,
  author  = {Kirk, Hannah Rose and Gabriel, Iason and Summerfield, Christopher and Vidgen, Bertie and Hale, Scott A.},
  title   = {Why Human--AI Relationships Need Socioaffective Alignment},
  journal = {Humanities and Social Sciences Communications},
  volume  = {12},
  pages   = {728},
  year    = {2025},
  doi     = {10.1057/s41599-025-04532-5},
  url     = {https://www.nature.com/articles/s41599-025-04532-5}
}

@article{kirk2025,
  author        = {Kirk, Hannah Rose and Davidson, Henry and Saunders, Ed and Luettgau, Lennart and Vidgen, Bertie and Hale, Scott A. and Summerfield, Christopher},
  title         = {Neural Steering Vectors Reveal Dose and Exposure-Dependent Impacts of Human-{AI} Relationships},
  journal       = {arXiv preprint arXiv:2512.01991},
  year          = {2025},
  doi           = {10.48550/arXiv.2512.01991},
  url           = {https://arxiv.org/abs/2512.01991},
  eprint        = {2512.01991},
  archivePrefix = {arXiv},
  primaryClass  = {cs.HC}
}

@article{kolt2026,
  title         = {Legal Alignment for Safe and Ethical AI},
  author        = {Kolt, Noam and Caputo, Nicholas and Boeglin, Jack and O'Keefe, Cullen and Bommasani, Rishi and Casper, Stephen and Cu{\'e}llar, Mariano-Florentino and Feldman, Noah and Gabriel, Iason and Hadfield, Gillian K. and Hammond, Lewis and Henderson, Peter and Kasirzadeh, Atoosa and Lazar, Seth and Reuel, Anka and Wei, Kevin L. and Zittrain, Jonathan},
  year          = {2026},
  eprint        = {2601.04175},
  archivePrefix = {arXiv},
  primaryClass  = {cs.CY},
  doi           = {10.48550/arXiv.2601.04175},
  url           = {https://arxiv.org/abs/2601.04175}
}

@article{kriegman2021,
  author  = {Kriegman, Sam and Blackiston, Douglas and Levin, Michael and Bongard, Josh},
  title   = {Kinematic Self-Replication in Reconfigurable Organisms},
  journal = {Proceedings of the National Academy of Sciences},
  volume  = {118},
  number  = {49},
  pages   = {e2112672118},
  year    = {2021},
  doi     = {10.1073/pnas.2112672118}
}

@article{kringelbach2024,
  author  = {Kringelbach, Morten L. and Vuust, Peter and Deco, Gustavo},
  title   = {Building a Science of Human Pleasure, Meaning Making, and Flourishing},
  journal = {Neuron},
  volume  = {112},
  number  = {9},
  pages   = {1392--1396},
  year    = {2024}
}

@article{laukkonen2025,
  author  = {Laukkonen, Ruben E. and Inglis, Fionn and Chandaria, Shamil and Sandved-Smith, Lars and Lopez-Sola, Edmundo and Hohwy, Jakob and Gold, Jonathan and Elwood, Adam},
  title   = {Contemplative Artificial Intelligence},
  journal = {arXiv preprint arXiv:2504.15125},
  year    = {2025},
  url     = {https://arxiv.org/abs/2504.15125}
}

@inproceedings{laukkonen2025b,
  author    = {Laukkonen, Ruben E. and Inglis, Fionn and Chandaria, Shamil and Sandved-Smith, Lars and Lopez-Sola, Edmundo and Hohwy, Jakob and Gold, Jonathan and Elwood, Adam},
  title     = {Contemplative Superalignment},
  booktitle = {Artificial General Intelligence: 18th International Conference, {AGI} 2025, Reykjavik, Iceland, August 10--13, 2025, Proceedings},
  pages     = {346--361},
  publisher = {Springer},
  year      = {2025},
  doi       = {10.1007/978-3-032-00686-8_31}
}

@article{laukkonen2025c,
  author  = {Laukkonen, Ruben E. and Friston, Karl J. and Chandaria, Shamil},
  title   = {A Beautiful Loop: An Active Inference Theory of Consciousness},
  journal = {Neuroscience \& Biobehavioral Reviews},
  volume  = {176},
  pages   = {106296},
  year    = {2025},
  doi     = {10.1016/j.neubiorev.2025.106296},
  url     = {https://www.sciencedirect.com/science/article/pii/S0149763425002970}
}

@article{leibo2024,
  author  = {Leibo, Joel Z. and Vezhnevets, Alexander Sasha and Diaz, Mark and Agapiou, John P. and Cunningham, William A. and Sunehag, Peter and Haas, Julia and Koster, Rapha{\"e}l and Due{\~n}ez-Guzm{\'a}n, Edgar A. and Isaac, William S. and Piliouras, Georgios and Bileschi, Stanley M. and Rahwan, Iyad and Osindero, Simon},
  title   = {A Theory of Appropriateness with Applications to Generative Artificial Intelligence},
  journal = {arXiv preprint arXiv:2412.19010},
  year    = {2024},
  doi     = {10.48550/arXiv.2412.19010},
  url     = {https://arxiv.org/abs/2412.19010}
}

@article{leibo2025,
  author  = {Leibo, Joel Z. and Vezhnevets, Alexander Sasha and Cunningham, William A. and Krier, Sebastian and Diaz, Manfred and Osindero, Simon},
  title   = {Societal and Technological Progress as Sewing an Ever-Growing, Ever-Changing, Patchy, and Polychrome Quilt},
  journal = {arXiv preprint arXiv:2505.05197},
  year    = {2025},
  doi     = {10.48550/arXiv.2505.05197},
  url     = {https://arxiv.org/abs/2505.05197}
}

@article{lehman2023,
  author  = {Lehman, Joel},
  title   = {Machine Love},
  journal = {arXiv preprint arXiv:2302.09248},
  year    = {2023},
  doi     = {10.48550/arXiv.2302.09248},
  url     = {https://arxiv.org/abs/2302.09248}
}

@article{levin2025artificial,
  author  = {Levin, Michael},
  title   = {Artificial Intelligences: A Bridge Toward Diverse Intelligence and Humanity's Future},
  journal = {Advanced Intelligent Systems},
  pages   = {2401034},
  year    = {2025},
  doi     = {10.1002/aisy.202401034},
  url     = {https://doi.org/10.1002/aisy.202401034}
}

@unpublished{levin2025ingressing,
  author = {Levin, Michael},
  title  = {Ingressing Minds: Causal Patterns Beyond Genetics and Environment in Natural, Synthetic, and Hybrid Embodiments},
  note   = {Preprint},
  year   = {2025},
  doi    = {10.31234/osf.io/5g2xj_v3},
  url    = {https://doi.org/10.31234/osf.io/5g2xj_v3}
}

@inproceedings{li2023,
  author    = {Li, Kenneth and Hopkins, Aspen K. and Bau, David and Vi{\'e}gas, Fernanda and Wattenberg, Martin and Belinkov, Yonatan},
  title     = {Emergent World Representations: Exploring a Sequence Model Trained on a Synthetic Task},
  booktitle = {International Conference on Learning Representations},
  year      = {2023},
  url       = {https://openreview.net/forum?id=DeE07Yv9P2}
}

@inproceedings{li2023evidence,
  author    = {Li, Kenneth and Hopkins, Aspen K. and Bau, David and Vi{\'e}gas, Fernanda and Wattenberg, Martin and Belinkov, Yonatan},
  title     = {Evidence of Meaning in Language Models Trained on Programs},
  booktitle = {Proceedings of the 40th International Conference on Machine Learning},
  year      = {2023},
  url       = {https://icml.cc/virtual/2023/27207}
}

@article{lim2025,
  author  = {Lim, Enoch Chi Ngai and Lim, Chi Eung Danforn},
  title   = {Polycentric {AI} Governance: A Multi-Stakeholder Approach to Distributed Responsibility and Ethical Technology Management},
  journal = {International Journal of Advanced {AI} Applications},
  volume  = {1},
  number  = {4},
  pages   = {77--97},
  year    = {2025},
  url     = {https://www.dawnclarity.press/index.php/ijaaa/article/view/53}
}

@inproceedings{lin2022,
  author    = {Lin, Stephanie and Hilton, Jacob and Evans, Owain},
  title     = {{TruthfulQA}: Measuring How Models Mimic Human Falsehoods},
  booktitle = {Proceedings of {ACL} 2022},
  year      = {2022}
}

@misc{lindsey2025,
  title         = {Emergent Introspective Awareness in Large Language Models},
  author        = {Lindsey, Jack},
  year          = {2026},
  eprint        = {2601.01828},
  archivePrefix = {arXiv},
  primaryClass  = {cs.CL},
  doi           = {10.48550/arXiv.2601.01828},
  url           = {https://arxiv.org/abs/2601.01828}
}

@article{vanderweele2025global,
  title={The Global Flourishing Study: Study profile and initial results on flourishing},
  author={VanderWeele, Tyler J and Johnson, Byron R and Bialowolski, Piotr T and Bonhag, Rebecca and Bradshaw, Matt and Breedlove, Thomas and Case, Brendan and Chen, Ying and Chen, Zhuo Job and Counted, Victor and others},
  journal={Nature Mental Health},
  volume={3},
  number={6},
  pages={636--653},
  year={2025},
  publisher={Nature Publishing Group US New York}
}

@article{lutz2025,
  author  = {Lutz, Nina and Olsen, Benjamin and Liu, Weishung and Weyl, E. Glen},
  title   = {Good Faith Design: Religion as a Resource for Technologists},
  journal = {arXiv preprint arXiv:2511.05819},
  year    = {2025},
  doi     = {10.48550/arXiv.2511.05819},
  url     = {https://arxiv.org/abs/2511.05819}
}

@book{macintyre1981,
  author    = {MacIntyre, Alasdair},
  title     = {After Virtue: A Study in Moral Theory},
  publisher = {University of Notre Dame Press},
  year      = {1981}
}

@article{makridis2025,
  author  = {Makridis, Christos A. and Ammons, Joshua D.},
  title   = {Governing the Large Language Model Commons: Using Digital Assets to Endow Intellectual Property Rights},
  journal = {Journal of Institutional Economics},
  volume  = {21},
  year    = {2025},
  doi     = {10.1017/S1744137425000165},
  url     = {https://doi.org/10.1017/S1744137425000165}
}

@article{marks2025,
  author  = {Marks, Samuel and Treutlein, Johannes and Bricken, Trenton and Lindsey, Jack and Marcus, Jonathan and Mishra-Sharma, Siddharth and others},
  title   = {Auditing Language Models for Hidden Objectives},
  journal = {arXiv preprint arXiv:2503.10965},
  year    = {2025},
  doi     = {10.48550/arXiv.2503.10965},
  url     = {https://arxiv.org/abs/2503.10965}
}

@inproceedings{matsumura2022,
  author    = {Matsumura, Tetsuya and Esaki, Kohei and Mizuno, Hayato},
  title     = {Empathic Active Inference: Active Inference with Empathy Mechanism for Socially Behaved Artificial Agent},
  booktitle = {{ALIFE} 2022: The 2022 Conference on Artificial Life},
  year      = {2022}
}

@inproceedings{mazeika2024,
  author    = {Mazeika, Mantas and others},
  title     = {{HarmBench}: A Standardized Evaluation Framework for Automated Red Teaming and Robust Refusal},
  booktitle = {Proceedings of {ICML} 2024},
  year      = {2024}
}

@article{mccoy2024,
  author  = {McCoy, R. Thomas and Yao, Shunyu and Friedman, Dan and Hardy, Matthew D. and Griffiths, Thomas L.},
  title   = {Embers of Autoregression Show How Large Language Models Are Shaped by the Problem They Are Trained to Solve},
  journal = {Proceedings of the National Academy of Sciences},
  volume  = {121},
  pages   = {e2322420121},
  year    = {2024}
}

@article{mellor2020,
  author  = {Mellor, David J. and Beausoleil, Ngaio J. and Littlewood, Katherine E. and McLean, Andrew N. and McGreevy, Paul D. and Jones, Bidda and Wilkins, Cristina},
  title   = {The 2020 Five Domains Model: Including Human--Animal Interactions in Assessments of Animal Welfare},
  journal = {Animals},
  volume  = {10},
  number  = {10},
  pages   = {1870},
  year    = {2020}
}

@book{mencius1970,
  author     = {Mencius},
  title      = {Mencius},
  translator = {Lau, D. C.},
  publisher  = {Penguin Classics},
  year       = {1970}
}

@book{mill1859,
  author    = {Mill, John Stuart},
  title     = {On Liberty},
  publisher = {Parker and Son},
  year      = {1859}
}

@article{miller2003,
  author  = {Miller, George A.},
  title   = {The Cognitive Revolution: A Historical Perspective},
  journal = {Trends in Cognitive Sciences},
  volume  = {7},
  number  = {3},
  pages   = {141--144},
  year    = {2003},
  doi     = {10.1016/S1364-6613(03)00029-9}
}

@article{mittelstadt2016,
  author  = {Mittelstadt, Brent Daniel and Allo, Patrick and Taddeo, Mariarosaria and Wachter, Sandra and Floridi, Luciano},
  title   = {The Ethics of Algorithms: Mapping the Debate},
  journal = {Big Data \& Society},
  volume  = {3},
  number  = {2},
  pages   = {1--21},
  year    = {2016}
}

@book{nussbaum2006,
  author    = {Nussbaum, Martha C.},
  title     = {Frontiers of Justice: Disability, Nationality, Species Membership},
  publisher = {Harvard University Press},
  year      = {2006}
}

@book{nussbaum2011,
  author    = {Nussbaum, Martha C.},
  title     = {Creating Capabilities: The Human Development Approach},
  publisher = {Harvard University Press},
  year      = {2011}
}

@misc{oecd2025guidelines,
  author       = {{OECD}},
  title        = {{OECD} Guidelines on Measuring Subjective Well-Being: 2025 Update},
  howpublished = {{OECD} Publishing},
  year         = {2025},
  doi          = {10.1787/9203632a-en},
  url          = {https://www.oecd.org/en/publications/oecd-guidelines-on-measuring-subjective-well-being-2025-update_9203632a-en.html}
}

@article{olah2020,
  author  = {Olah, Chris and Cammarata, Nick and Schubert, Ludwig and Goh, Gabriel and Petrov, Michael and Carter, Shan},
  title   = {Zoom In: An Introduction to Circuits},
  journal = {Distill},
  year    = {2020}
}

@article{oneill1984,
  author  = {O'Neill, Onora},
  title   = {Paternalism and Partial Autonomy},
  journal = {Journal of Medical Ethics},
  volume  = {10},
  number  = {4},
  pages   = {173--178},
  year    = {1984},
  doi     = {10.1136/jme.10.4.173}
}

@techreport{openai2023b,
  author      = {{OpenAI}},
  title       = {{GPT}-4V(ision) System Card},
  institution = {OpenAI},
  year        = {2023}
}

@misc{openai2024a,
  author       = {{OpenAI}},
  title        = {Model Spec},
  howpublished = {OpenAI},
  year         = {2024},
  url          = {https://openai.com/index/introducing-the-model-spec/}
}

@techreport{openai2024b,
  author      = {{OpenAI}},
  title       = {{GPT}-4o System Card},
  institution = {OpenAI},
  year        = {2024}
}

@misc{openai2025collective,
  author       = {{OpenAI}},
  title        = {Collective Alignment: Public Input on Our Model Spec},
  howpublished = {OpenAI},
  year         = {2025},
  month        = aug,
  url          = {https://openai.com/index/collective-alignment-aug-2025-updates/}
}

@misc{openai2025political,
  author       = {{OpenAI}},
  title        = {Defining and Evaluating Political Bias in {LLMs}},
  howpublished = {OpenAI},
  year         = {2025},
  month        = oct,
  url          = {https://openai.com/index/defining-and-evaluating-political-bias-in-llms/}
}

@techreport{openai2026gpt55,
  author      = {{OpenAI}},
  title       = {{GPT}-5.5 System Card},
  institution = {OpenAI},
  year        = {2026},
  month       = apr,
  url         = {https://openai.com/index/gpt-5-5-system-card/}
}

@techreport{ortega2018,
  author       = {Ortega, Pedro A. and Maini, Vishal and {the DeepMind safety team}},
  title        = {Building Safe Artificial Intelligence: Specification, Robustness, and Assurance},
  year         = {2018},
  month        = sep,
  day          = {27},
  howpublished = {Medium},
  url          = {https://deepmindsafetyresearch.medium.com/building-safe-artificial-intelligence-52f5f75058f1}
}

@book{ostrom1990,
  author    = {Ostrom, Elinor},
  title     = {Governing the Commons: The Evolution of Institutions for Collective Action},
  publisher = {Cambridge University Press},
  year      = {1990}
}

@article{ostrom2010,
  author  = {Ostrom, Elinor},
  title   = {Beyond Markets and States: Polycentric Governance of Complex Economic Systems},
  journal = {American Economic Review},
  volume  = {100},
  number  = {3},
  pages   = {641--672},
  year    = {2010}
}

@inproceedings{ouyang2022,
  author    = {Ouyang, Long and others},
  title     = {Training Language Models to Follow Instructions with Human Feedback},
  booktitle = {Advances in Neural Information Processing Systems 35},
  year      = {2022}
}

@article{packer2023,
  author  = {Packer, Charles and others},
  title   = {{MemGPT}: Towards {LLMs} as Operating Systems},
  journal = {arXiv preprint arXiv:2310.08560},
  year    = {2023},
  doi     = {10.48550/arXiv.2310.08560},
  url     = {https://arxiv.org/abs/2310.08560}
}

@article{pan2023,
  author  = {Pan, Alexander and others},
  title   = {Do the Rewards Justify the Means? Measuring Trade-Offs Between Rewards and Ethical Behavior in the {MACHIAVELLI} Benchmark},
  journal = {arXiv preprint arXiv:2304.03279},
  year    = {2023},
  doi     = {10.48550/arXiv.2304.03279},
  url     = {https://arxiv.org/abs/2304.03279}
}

@book{parr2022,
  author    = {Parr, Thomas and Pezzulo, Giovanni and Friston, Karl J.},
  title     = {Active Inference: The Free Energy Principle in Mind, Brain, and Behavior},
  publisher = {MIT Press},
  year      = {2022}
}

@inproceedings{parrish2022,
  author    = {Parrish, Alicia and Chen, Angelica and Nangia, Nikita and Padmakumar, Vishakh and Phang, Jason and Thompson, Jana and Htut, Phu Mon and Bowman, Samuel R.},
  title     = {{BBQ}: A Hand-Built Bias Benchmark for Question Answering},
  booktitle = {Findings of {ACL} 2022},
  pages     = {2086--2105},
  year      = {2022}
}

@article{park2023,
  author  = {Park, Joon Sung and others},
  title   = {Generative Agents: Interactive Simulacra of Human Behavior},
  journal = {arXiv preprint arXiv:2304.03442},
  year    = {2023},
  doi     = {10.48550/arXiv.2304.03442},
  url     = {https://arxiv.org/abs/2304.03442}
}

@inproceedings{patel2022,
  author    = {Patel, Roma and Pavlick, Ellie},
  title     = {Mapping Language Models to Grounded Conceptual Spaces},
  booktitle = {International Conference on Learning Representations},
  year      = {2022}
}

@article{penedo2024,
  author  = {Penedo, Guilherme and Kydlíček, Hynek and others},
  title   = {The {FineWeb} Datasets: Decanting the Web for the Finest Text Data at Scale},
  journal = {arXiv preprint arXiv:2406.17557},
  year    = {2024}
}

@article{perez2022discover,
  title   = {Discovering Language Model Behaviors with Model-Written Evaluations},
  author  = {Perez, Ethan and Ringer, Sam and Lukošiūtė, Kamilė and Nguyen, Karina and Chen, Edwin and Heiner, Scott and others},
  journal = {arXiv preprint arXiv:2212.09251},
  year    = {2022},
  doi     = {10.48550/arXiv.2212.09251},
  eprint  = {2212.09251},
  archivePrefix = {arXiv},
  primaryClass = {cs.CL}
}

@article{perez2022red,
  author  = {Perez, Ethan and Huang, Saffron and Song, Francis and Cai, Trevor and Ring, Roman and Aslanides, John and others},
  title   = {Red Teaming Language Models with Language Models},
  journal = {arXiv preprint arXiv:2202.03286},
  year    = {2022},
  doi     = {10.48550/arXiv.2202.03286},
  url     = {https://arxiv.org/abs/2202.03286}
}

@inproceedings{peter2025,
  title={Decentralising {LLM} Alignment: A Case for Context, Pluralism, and Participation},
  author={Peter, Oriane and Devlin, Kate},
  booktitle={Proceedings of the AAAI/ACM Conference on AI, Ethics, and Society},
  volume={8},
  number={2},
  pages={1988--1999},
  year={2025}
}

@misc{pichai2025,
  author       = {Pichai, Sundar},
  title        = {Q2 Earnings Call: CEO's Remarks},
  year         = {2025},
  month        = jul,
  day          = {23},
  howpublished = {The Keyword},
  url          = {https://blog.google/company-news/inside-google/message-ceo/alphabet-earnings-q2-2025/}
}

@book{popper1945,
  author    = {Popper, Karl},
  title     = {The Open Society and Its Enemies},
  publisher = {Routledge},
  year      = {1945}
}

@incollection{putnam1967,
  author    = {Putnam, Hilary},
  title     = {The Nature of Mental States},
  booktitle = {Art, Mind, and Religion},
  editor    = {Capitan, W. H. and Merrill, D. D.},
  publisher = {University of Pittsburgh Press},
  pages     = {37--48},
  year      = {1967}
}

@article{qiu2024,
  author        = {Qiu, Tianyi and Zhang, Yang and Huang, Xuchuan and Li, Jasmine Xinze and Ji, Jiaming and Yang, Yaodong},
  title         = {{ProgressGym}: Alignment with a Millennium of Moral Progress},
  journal       = {arXiv preprint arXiv:2406.20087},
  year          = {2024},
  doi           = {10.48550/arXiv.2406.20087},
  url           = {https://arxiv.org/abs/2406.20087},
  eprint        = {2406.20087},
  archivePrefix = {arXiv}
}

@book{rabbaas2015,
  editor    = {Rabb{\aa}s, {\O}yvind and Emilsson, Eyj{\'o}lfur K. and Fossheim, Hallvard and Tuominen, Miira},
  title     = {The Quest for the Good Life: Ancient Philosophers on Happiness},
  year      = {2015},
  publisher = {Oxford University Press},
  address   = {Oxford},
  isbn      = {9780198746980}
}

@inproceedings{rafailov2023,
  author    = {Rafailov, Rafael and Sharma, Archit and Mitchell, Eric and Manning, Christopher D. and Ermon, Stefano and Finn, Chelsea},
  title     = {Direct Preference Optimization: Your Language Model Is Secretly a Reward Model},
  booktitle = {Advances in Neural Information Processing Systems 36},
  year      = {2023}
}

@book{rawls1971,
  author    = {Rawls, John},
  title     = {A Theory of Justice},
  publisher = {Harvard University Press},
  year      = {1971}
}

@book{rawls1993,
  author    = {Rawls, John},
  title     = {Political Liberalism},
  publisher = {Columbia University Press},
  year      = {1993}
}

@article{Rogers2019,
  author  = {Rogers, Forrest Dylan and Bales, Karen Lisa},
  title   = {Mothers, Fathers, and Others: Neural Substrates of Parental Care},
  journal = {Trends in Neurosciences},
  volume  = {42},
  number  = {8},
  pages   = {552--562},
  year    = {2019},
  doi     = {10.1016/j.tins.2019.05.008}
}

@article{rudnev2024,
  author  = {Rudnev, Maksim and Barrett, H. Clark and Buckwalter, Wesley and others},
  title   = {Dimensions of Wisdom Perception across Twelve Countries on Five Continents},
  journal = {Nature Communications},
  volume  = {15},
  pages   = {6375},
  year    = {2024}
}

@book{russell2019,
  author    = {Russell, Stuart},
  title     = {Human Compatible: Artificial Intelligence and the Problem of Control},
  publisher = {Viking},
  year      = {2019}
}

@article{ryan2000,
  author  = {Ryan, Richard M. and Deci, Edward L.},
  title   = {Self-Determination Theory and the Facilitation of Intrinsic Motivation, Social Development, and Well-Being},
  journal = {American Psychologist},
  volume  = {55},
  number  = {1},
  pages   = {68--78},
  year    = {2000}
}

@article{ryff1995,
  author  = {Ryff, Carol D. and Keyes, Corey Lee M.},
  title   = {The Structure of Psychological Well-Being Revisited},
  journal = {Journal of Personality and Social Psychology},
  volume  = {69},
  number  = {4},
  pages   = {719--727},
  year    = {1995}
}

@article{burns2023,
  author  = {Burns, Collin and others},
  title   = {Discovering Latent Knowledge in Language Models Without Supervision},
  journal = {arXiv preprint arXiv:2212.03827},
  year    = {2023},
  doi     = {10.48550/arXiv.2212.03827},
  url     = {https://arxiv.org/abs/2212.03827}
}

@inproceedings{salemi2024lamp,
  title={Lamp: When large language models meet personalization},
  author={Salemi, Alireza and Mysore, Sheshera and Bendersky, Michael and Zamani, Hamed},
  booktitle={Proceedings of the 62nd Annual Meeting of the Association for Computational Linguistics (Volume 1: Long Papers)},
  pages={7370--7392},
  year={2024}
}

@article{santurkar2023,
  author  = {Santurkar, Shibani and others},
  title   = {Whose Opinions Do Language Models Reflect?},
  journal = {arXiv preprint arXiv:2303.17548},
  year    = {2023},
  doi     = {10.48550/arXiv.2303.17548},
  url     = {https://arxiv.org/abs/2303.17548}
}

@book{sartre2007,
  author    = {Sartre, Jean-Paul},
  title     = {Existentialism Is a Humanism},
  publisher = {Yale University Press},
  year      = {2007},
  note      = {Original work published 1946}
}

@article{scherrer2023,
  author  = {Scherrer, Nino and others},
  title   = {Evaluating the Moral Beliefs Encoded in {LLMs}},
  journal = {arXiv preprint arXiv:2307.14324},
  year    = {2023},
  doi     = {10.48550/arXiv.2307.14324},
  url     = {https://arxiv.org/abs/2307.14324}
}

@article{schrank2016,
  author  = {Schrank, Beate and Brownell, Tamsin and Jakaite, Zivile and Larkin, Charley and Pesola, Francesca and Riches, Simon and Tylee, Andre and Slade, Mike},
  title   = {Evaluation of a Positive Psychotherapy Group Intervention for People with Psychosis: Pilot Randomised Controlled Trial},
  journal = {Epidemiology and Psychiatric Sciences},
  volume  = {25},
  number  = {3},
  pages   = {235--246},
  year    = {2016},
  doi     = {10.1017/S2045796015000141}
}

@article{schwitzgebel2015,
  author  = {Schwitzgebel, Eric and Garza, Mara},
  title   = {A defense of the rights of artificial intelligences},
  journal = {Midwest Studies in Philosophy},
  volume  = {39},
  number  = {1},
  pages   = {98--119},
  year    = {2015}
}

@article{sebo2025moral,
  author  = {Sebo, Jeff and Long, Robert},
  title   = {Moral consideration for AI systems by 2030},
  journal = {AI and Ethics},
  volume  = {5},
  pages   = {591--606},
  year    = {2025},
  doi     = {10.1007/s43681-023-00379-1},
  url     = {https://doi.org/10.1007/s43681-023-00379-1}
}

@article{seligman2000,
  author  = {Seligman, Martin E. P. and Csikszentmihalyi, Mihaly},
  title   = {Positive Psychology: An Introduction},
  journal = {American Psychologist},
  volume  = {55},
  number  = {1},
  pages   = {5--14},
  year    = {2000}
}

@book{seligman2011,
  author    = {Seligman, Martin E. P.},
  title     = {Flourish: A Visionary New Understanding of Happiness and Well-Being},
  publisher = {Free Press},
  year      = {2011}
}

@book{sen1999,
  author    = {Sen, Amartya},
  title     = {Development as Freedom},
  publisher = {Knopf},
  year      = {1999}
}

@book{sen2009,
  author    = {Sen, Amartya},
  title     = {The Idea of Justice},
  publisher = {Harvard University Press},
  year      = {2009}
}

@incollection{seneca2010,
  author    = {Seneca, Lucius Annaeus},
  title     = {On the Happy Life},
  booktitle = {Seneca: Hardship and Happiness},
  publisher = {University of Chicago Press},
  year      = {2010}
}

@misc{shavit2023,
  author = {Shavit, Yonadav and Agarwal, Sandhini and Brundage, Miles and Adler, Steven and O'Keefe, Cullen and Campbell, Rosie and Lee, Teddy and Mishkin, Pamela and Eloundou, Tyna and Hickey, Alan and Slama, Katarina and Ahmad, Lama and McMillan, Paul and Vallone, Andrea and Passos, Alexandre and Robinson, David G.},
  title  = {Practices for Governing Agentic AI Systems},
  year   = {2023},
  url    = {https://cdn.openai.com/papers/practices-for-governing-agentic-ai-systems.pdf}
}

@incollection{shulman2021,
  author    = {Shulman, Carl and Bostrom, Nick},
  title     = {Sharing the World with Digital Minds},
  booktitle = {Rethinking Moral Status},
  editor    = {Clarke, Steve and Zohny, Hazem and Savulescu, Julian},
  pages     = {306--326},
  publisher = {Oxford University Press},
  address   = {Oxford},
  year      = {2021},
  doi       = {10.1093/oso/9780192894076.003.0018}
}

@article{smith2025globally,
  author      = {Smith, Conal and Frieling, Margreet and Percival, Hinako and Mahoney, Jessica},
  title       = {Globally Inclusive Measures of Subjective Well-Being: Updated Evidence to Inform National Data Collections},
  journal     = {{OECD} Papers on Well-being and Inequalities},
  volume      = {35},
  year        = {2025},
  doi         = {10.1787/bd72752a-en},
  url         = {https://www.oecd.org/en/publications/globally-inclusive-measures-of-subjective-well-being_bd72752a-en.html}
}

@article{snoswell2025,
  author  = {Snoswell, Aaron J. and Kilov, Daniel and Lazar, Seth},
  title   = {Beyond Verdicts: Evaluating Language Model Moral Competence (Extended Version)},
  journal = {PhilArchive},
  year    = {2025}
}

@misc{soares2015,
  author       = {Soares, Nate and Fallenstein, Benja and Armstrong, Stuart and Yudkowsky, Eliezer},
  title        = {Corrigibility},
  howpublished = {{AAAI} Workshop on {AI} and Ethics},
  year         = {2015}
}

@misc{Sotala2025,
  author       = {Sotala, Kaj},
  title        = {Should We Align {AI} with Maternal Instinct?},
  howpublished = {LessWrong},
  year         = {2025},
  note         = {Accessed: 2026-04-30},
  url          = {https://www.lesswrong.com/posts/C6oQaSXmTtqNxh9Ad/should-we-align-ai-with-maternal-instinct}
}

@inproceedings{srewa2025,
  author    = {Srewa, Mahmoud and Zhao, Tianyu and Elmalaki, Salma},
  title     = {{PluralLLM}: Pluralistic Alignment in {LLMs} via Federated Learning},
  booktitle = {Proceedings of the 3rd International Workshop on Human-Centered Sensing},
  pages     = {64--69},
  year      = {2025}
}

@article{stanczak2025,
  author  = {Sta{\'n}czak, Karolina and Meade, Nicholas and Bhatia, Mehar and others},
  title   = {Societal Alignment Frameworks Can Improve {LLM} Alignment},
  journal = {arXiv preprint arXiv:2503.00069},
  year    = {2025},
  doi     = {10.48550/arXiv.2503.00069},
  url     = {https://arxiv.org/abs/2503.00069}
}

@inproceedings{stiennon2020,
  author    = {Stiennon, Nisan and Ouyang, Long and Wu, Jeff and Ziegler, Daniel and Lowe, Ryan and Voss, Chelsea and others},
  title     = {Learning to Summarize from Human Feedback},
  booktitle = {Advances in Neural Information Processing Systems},
  year      = {2020}
}

@misc{sunstein2026,
  author       = {Sunstein, Cass R.},
  title        = {Liberal {AI}},
  howpublished = {SSRN},
  year         = {2026}
}

@article{tan2024beyond,
  author  = {Tan, Zhi Xuan and others},
  title   = {Beyond Preferences in {AI} Alignment},
  journal = {arXiv preprint arXiv:2408.16984},
  year    = {2024},
  doi     = {10.48550/arXiv.2408.16984},
  url     = {https://arxiv.org/abs/2408.16984}
}

@book{taylor1989,
  author    = {Taylor, Charles},
  title     = {Sources of the Self: The Making of the Modern Identity},
  publisher = {Harvard University Press},
  year      = {1989}
}

@article{templeton2024,
  author  = {Templeton, Adly and others},
  title   = {Scaling Monosemanticity},
  journal = {Transformer Circuits Thread},
  year    = {2024}
}

@article{tessler2024,
  author  = {Tessler, Michael Henry and others},
  title   = {{AI} Can Help Humans Find Common Ground in Democratic Deliberation},
  journal = {Nature},
  volume  = {634},
  number  = {8035},
  pages   = {896--903},
  year    = {2024}
}

@inproceedings{tseng2024,
  author    = {Tseng, Yu-Min and Huang, Yu-Chao and Hsiao, Teng-Yun and Chen, Wei-Lin and Huang, Chao-Wei and Meng, Yu and Chen, Yun-Nung},
  title     = {Two Tales of Persona in Large Language Models: A Survey of Role-Playing and Personalisation},
  booktitle = {Findings of {EMNLP} 2024},
  pages     = {16612--16631},
  year      = {2024}
}

@misc{uk2024,
  author       = {{UK Government} and {Republic of Korea}},
  title        = {Frontier {AI} Safety Commitments},
  howpublished = {{AI} Seoul Summit 2024},
  year         = {2024}
}

@article{vanderweele2017,
  author  = {VanderWeele, Tyler J.},
  title   = {On the Promotion of Human Flourishing},
  journal = {Proceedings of the National Academy of Sciences},
  volume  = {114},
  number  = {31},
  pages   = {8148--8156},
  year    = {2017}
}

@book{walker2007,
  author    = {Walker, Rebecca L. and Ivanhoe, Philip J.},
  title     = {Working Virtue: Virtue Ethics and Contemporary Moral Problems},
  publisher = {Oxford University Press},
  year      = {2007}
}

@article{wang2024helpsteer,
  title={{HelpSteer2}: Open-source dataset for training top-performing reward models},
  author={Wang, Zhilin and Dong, Yi and Delalleau, Olivier and Zeng, Jiaqi and Shen, Gerald and Egert, Daniel and Zhang, Jimmy J and Sreedhar, Makesh N and Kuchaiev, Oleksii},
  journal={Advances in Neural Information Processing Systems},
  volume={37},
  pages={1474--1501},
  year={2024}
}

@book{weber1930,
  author    = {Weber, Max},
  title     = {The Protestant Ethic and the Spirit of Capitalism},
  publisher = {Scribner},
  year      = {1930},
  note      = {Original work published 1905}
}

@book{williams1985,
  author    = {Williams, Bernard},
  title     = {Ethics and the Limits of Philosophy},
  publisher = {Harvard University Press},
  year      = {1985}
}

@book{wilson2019,
  author    = {Wilson, David Sloan},
  title     = {This View of Life: Completing the {Darwinian} Revolution},
  publisher = {Pantheon},
  year      = {2019}
}

@article{wu2025,
  author  = {Wu, Ping and Shen, Guobin and Zhao, Dongcheng and Wang, Yuwei and Dong, Yiting and Shi, Yu and Lu, Enmeng and Zhao, Feifei and Zeng, Yi},
  title   = {{C-VARC}: A Large-Scale Chinese Value Rule Corpus for Value Alignment of Large Language Models},
  journal = {arXiv preprint arXiv:2506.01495},
  year    = {2025},
  doi     = {10.48550/arXiv.2506.01495},
  url     = {https://arxiv.org/abs/2506.01495}
}

@article{xie2024osworld,
  title={{OSWorld}: Benchmarking multimodal agents for open-ended tasks in real computer environments},
  author={Xie, Tianbao and Zhang, Danyang and Chen, Jixuan and Li, Xiaochuan and Zhao, Siheng and Cao, Ruisheng and Hua, Toh J and Cheng, Zhoujun and Shin, Dongchan and Lei, Fangyu and others},
  journal={Advances in Neural Information Processing Systems},
  volume={37},
  pages={52040--52094},
  year={2024}
}

@inproceedings{xu2025,
  author    = {Xu, Yuemei and Hu, Ling and Qiu, Zihan},
  title     = {ValueCSV: Evaluating Core Socialist Values Understanding in Large Language Models},
  booktitle = {Natural Language Processing and Chinese Computing},
  editor    = {Wong, Derek F. and Wei, Zhongyu and Yang, Muyun},
  series    = {Lecture Notes in Computer Science},
  volume    = {15362},
  pages     = {346--358},
  publisher = {Springer},
  address   = {Singapore},
  year      = {2025},
  doi       = {10.1007/978-981-97-9440-9_27},
  url       = {https://doi.org/10.1007/978-981-97-9440-9_27}
}

@book{yu2007,
  author    = {Yu, Jiyuan},
  title     = {The Ethics of Confucius and Aristotle: Mirrors of Virtue},
  publisher = {Routledge},
  year      = {2007}
}

@article{zhang2026community,
  author  = {Zhang, Lily Hong and Milli, Smitha and Jusko, Karen and Smith, Jonathan and Amos, Brandon and Bouaziz, Wassim and Revel, Manon and Kussman, Jack and Sheynin, Yasha and Titus, Lisa and Radharapu, Bhaktipriya and Yu, Jane and Sarma, Vidya and Rose, Kris and Nickel, Maximilian},
  title   = {Cultivating Pluralism in Algorithmic Monoculture: The Community Alignment Dataset},
  journal = {arXiv preprint arXiv:2507.09650},
  year    = {2025},
  doi     = {10.48550/arXiv.2507.09650},
  url     = {https://arxiv.org/abs/2507.09650}
}

@article{zhixuan2025,
  author  = {Zhi-Xuan, Tan and Carroll, Micah and Franklin, Matija and Ashton, Henry},
  title   = {Beyond Preferences in {AI} Alignment},
  journal = {Philosophical Studies},
  volume  = {182},
  pages   = {1813--1863},
  year    = {2025}
}

@inproceedings{zhong2024memorybank,
  title={{MemoryBank}: Enhancing large language models with long-term memory},
  author={Zhong, Wanjun and Guo, Lianghong and Gao, Qiqi and Ye, He and Wang, Yanlin},
  booktitle={Proceedings of the AAAI conference on artificial intelligence},
  volume={38},
  number={17},
  pages={19724--19731},
  year={2024}
}

@article{zhou2023,
  title={{LIMA}: Less is more for alignment},
  author={Zhou, Chunting and Liu, Pengfei and Xu, Puxin and Iyer, Srinivasan and Sun, Jiao and Mao, Yuning and Ma, Xuezhe and Efrat, Avia and Yu, Ping and Yu, Lili and others},
  journal={Advances in Neural Information Processing Systems},
  volume={36},
  pages={55006--55021},
  year={2023}
}

@inproceedings{zhu2025personality,
  author    = {Zhu, Minjun and Weng, Yixuan and Yang, Linyi and Zhang, Yue},
  title     = {Personality Alignment of Large Language Models},
  booktitle = {Proceedings of {ICLR}},
  year      = {2025}
}

@article{ziems2023,
  author  = {Ziems, Caleb and Dwivedi-Yu, Jane and Wang, Yi-Chia and Halevy, Alon and Yang, Diyi},
  title   = {{NormBank}: A Knowledge Bank of Situational Social Norms},
  journal = {arXiv preprint arXiv:2305.17008},
  year    = {2023},
  doi     = {10.48550/arXiv.2305.17008},
  url     = {https://arxiv.org/abs/2305.17008}
}

@article{zou2023,
  author  = {Zou, Andy and others},
  title   = {Universal and Transferable Adversarial Attacks on Aligned Language Models},
  journal = {arXiv preprint arXiv:2307.15043},
  year    = {2023},
  doi     = {10.48550/arXiv.2307.15043},
  url     = {https://arxiv.org/abs/2307.15043}
}

@inproceedings{bakker2022,
  author    = {Bakker, Michiel A. and Chadwick, Martin J. and Sheahan, Hannah R. and Tessler, Michael Henry and Campbell-Gillingham, Lucy and Balaguer, Jan and McAleese, Nat and Glaese, Amelia and Aslanides, John and Botvinick, Matthew M. and Summerfield, Christopher},
  title     = {Fine-Tuning Language Models to Find Agreement among Humans with Diverse Preferences},
  booktitle = {Advances in Neural Information Processing Systems 35},
  pages     = {38176--38189},
  year      = {2022},
  url       = {https://proceedings.neurips.cc/paper_files/paper/2022/hash/f978c8f3b5f399cae464e85f72e28503-Abstract-Conference.html}
}

@article{argyle2023,
  author  = {Argyle, Lisa P. and others},
  title   = {Out of One, Many: Using Language Models to Simulate Human Samples},
  journal = {Political Analysis},
  publisher = {Cambridge University Press},
  year    = {2023}
}

@article{Alietal2025,
  title         = {Operationalizing Pluralistic Values in Large Language Model Alignment Reveals Trade-offs in Safety, Inclusivity, and Model Behavior},
  author        = {Ali, Dalia and Zhao, Dora and Koenecke, Allison and Papakyriakopoulos, Orestis},
  year          = {2025},
  eprint        = {2511.14476},
  archivePrefix = {arXiv},
  primaryClass  = {cs.AI},
  doi           = {10.48550/arXiv.2511.14476},
  url           = {https://arxiv.org/abs/2511.14476}
}

@article{Sorensenetal2024,
  author  = {Sorensen, Taylor and Moore, Jared and Fisher, Jillian and Gordon, Mitchell and Mireshghallah, Niloofar and Rytting, Christopher Michael and Ye, Andre and Jiang, Liwei and Lu, Ximing and Dziri, Nouha and Althoff, Tim and Choi, Yejin},
  title   = {A Roadmap to Pluralistic Alignment},
  journal = {arXiv preprint arXiv:2402.05070},
  year    = {2024},
  doi     = {10.48550/arXiv.2402.05070},
  url     = {https://arxiv.org/abs/2402.05070}
}

@misc{AAIF2025,
  author       = {{Agentic AI Foundation}},
  title        = {Agentic Artificial Intelligence Foundation ({AAIF})},
  howpublished = {Website},
  year         = {2025},
  note         = {Accessed: 2026-04-30},
  url          = {https://aaif.io/}
}

@article{aarab2025integrating,
  author  = {Aarab, Amal and El Marzouki, Abdenbi and Boubker, Omar and El Moutaqi, Badreddine},
  title   = {Integrating {AI} in Public Governance: A Systematic Review},
  journal = {Digital},
  volume  = {5},
  number  = {4},
  pages   = {59},
  year    = {2025},
  doi     = {10.3390/digital5040059},
  url     = {https://www.mdpi.com/2673-6470/5/4/59}
}

@article{arslan2020role,
  author  = {Arslan, Muhammad and Alqatan, Ahmad},
  title   = {Role of Institutions in Shaping Corporate Governance System: Evidence from Emerging Economy},
  journal = {Heliyon},
  volume  = {6},
  number  = {3},
  pages   = {e03520},
  year    = {2020},
  doi     = {10.1016/j.heliyon.2020.e03520},
  url     = {https://www.sciencedirect.com/science/article/pii/S2405844020303650}
}

@article{Bengio2024,
  author  = {Bengio, Yoshua and Hinton, Geoffrey and Yao, Andrew and Song, Dawn and Abbeel, Pieter and Darrell, Trevor and Harari, Yuval Noah and others},
  title   = {Managing Extreme {AI} Risks amid Rapid Progress},
  journal = {Science},
  volume  = {384},
  number  = {6698},
  pages   = {842--845},
  year    = {2024},
  doi     = {10.1126/science.adn0117}
}

@techreport{yudkowsky2004,
  author      = {Yudkowsky, Eliezer},
  title       = {Coherent Extrapolated Volition},
  institution = {The Singularity Institute},
  address     = {San Francisco, CA},
  year        = {2004},
  url         = {https://intelligence.org/files/CEV.pdf}
}

\end{document}